\documentclass{article}
\usepackage{frExamplee}
\usepackage{graphicx}
\usepackage{apalike}
\usepackage{setspace}

\usepackage{graphicx} 
\usepackage[table, dvipsnames]{xcolor}
\usepackage{amsmath} 
\usepackage{amssymb}  
\usepackage{acronym}
\usepackage{todonotes}
\usepackage[font=small]{caption}
\usepackage{subcaption}
\usepackage{array,booktabs}
\usepackage{url}
\usepackage{hyperref}
\usepackage[ruled,vlined]{algorithm2e}
\usepackage[noabbrev]{cleveref}
\usepackage{pbox}
\usepackage{scalerel}
\usepackage{siunitx}

\graphicspath{{fig/}}

\newcommand{\edited}[1]{#1}

\newcommand{\mynumber}[1]{\protect\scalerel*{\protect\includegraphics{fig/numbers/#1}}{#1}}

\acrodef{RNN}{rapidly exploring random tree}
\acrodef{RL}{reinforcement learning}
\acrodef{DoF}{degrees of freedom}
\acrodef{TSDF}{truncated signed-distance field}
\acrodef{CNN}{convolutional neural network}
\acrodef{SubT}{DARPA Subterranean Challenge}
\acrodef{DARPA}{Defense Advanced Research Projects Agency}
\acrodef{FoV}{field-of-view}

\title{
ArtPlanner:\\Robust Legged Robot Navigation in the Field
}

\author{Lorenz Wellhausen  \\
Robotic Systems Lab \\ ETH Z{\"u}rich \\
Switzerland \\

\And

Marco Hutter \\
Robotic Systems Lab \\ ETH Z{\"u}rich \\
Switzerland%
\thanks{This work was supported by the Swiss National Science Foundation
(SNF) through the NCCR Robotics and project 188596, and the EU Horizon 2020 research and innovation programme under grant agreements No 780883, No 852044 and No 101016970. 
It has been conducted as part of ANYmal Research, a community to advance legged robotics}

}

\begin{document}
\maketitle


\begin{abstract}
Due to the highly complex environment present during the DARPA Subterranean Challenge, all six funded teams relied on legged robots as part of their robotic team.
Their unique locomotion skills of being able to step over obstacles require special considerations for navigation planning.
In this work, we present and examine ArtPlanner, the navigation planner used by team CERBERUS during the Finals.
It is based on a sampling-based method that determines valid poses with a reachability abstraction and uses learned foothold scores to restrict areas considered safe for stepping.
The resulting planning graph is assigned learned motion costs by a neural network trained in simulation to minimize traversal time and limit the risk of failure.
Our method\footnote{\url{http://github.com/leggedrobotics/art_planner}} achieves real-time performance with a bounded computation time.
We present extensive experimental results gathered during the Finals event of the DARPA Subterranean Challenge, where this method contributed to team CERBERUS winning the competition.
It powered navigation of four ANYmal quadrupeds for 90 minutes of autonomous operation without a single \edited{planning or locomotion} failure.
\end{abstract}


\section{Introduction}
\label{sec:introduction}

\begin{figure}[t]
  \includegraphics[width=\linewidth]{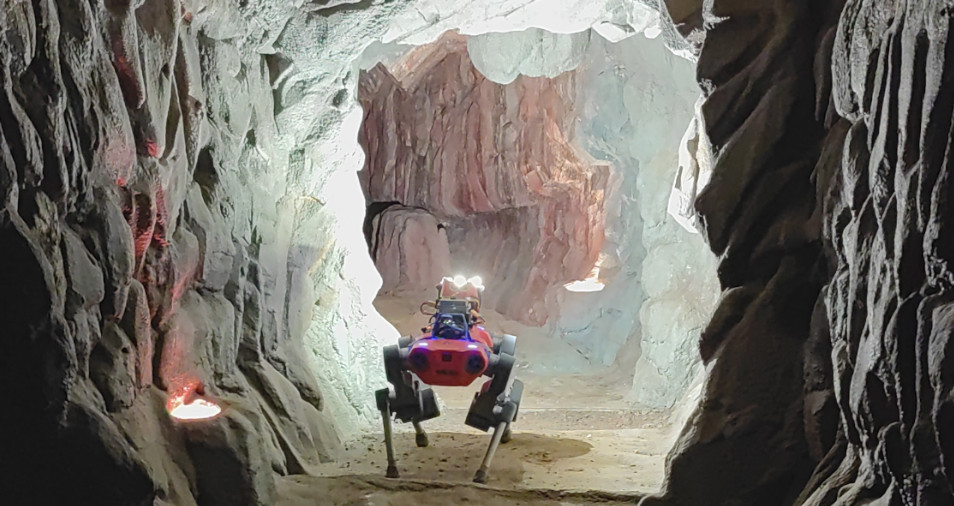}
  \caption{Team CERBERUS won the DARPA Subterranean Challenge Finals with four ANYmal quadrupeds deployed during the Prize Run. The navigation planner presented in this work guided all four robots safely during the hour-long mission.}
  \label{fig:teaser}
\end{figure}

Navigation planning for legged robots has distinct challenges which are not present for other types of robots.
While flying robots attempt to avoid any contact with the environment, ground robots by definition require contact with the ground to locomote. 
Compared to other types of ground robots, which have a constant contact patch with the ground, legged robots can overcome obstacles by lifting their legs. 
Most traditional navigation planning approaches assume a single traversability value for any given terrain patch, which they check against the footprint of the robot~\cite{wermelinger2016navigation}.
These approaches are limiting for legged robots because their ability to change their footprint and choose contact locations with the environment deliberately is not accounted for.
Defining a traversability value for such a highly articulated systems is extremely challenging due to the high dimensionality of the problem.

To find a stable configuration for a legged robot, we need to find a base pose where the terrain is within the range of motion of each limb.
We call this reachability checking, where a robot is represented as one collision volume for its torso, and one reachability volume for each of its limbs~\cite{tonneau2015reachability}.
When checking the feasability of a given robot pose, the torso volume is expected to be collision-free, while collision is enforced for the reachability volumes, to ensure that the robot is able to make environment-contact with its legs.
While this approach can find all theoretically feasible poses, basic reachability does not account for dynamics of the robot and in practice most locomotion controllers do not use the full range of motion.
This issue can somewhat be alleviated by restricting geometry which is considered safe for contacts, and heuristics on the range of motion used by a specific locomotion controller.
This mostly produces feasible paths, but the cost of walking over challenging terrain and the risk of stepping close to obstacles and edges is not considered~\cite{wellhausen2021rough}.

In this work, we consider the path planning problem for legged robots on a robo-centric height map, which is of fixed size, centered around the robot.
Since we are interested in a navigation planner which can work in possibly unknown environments, it will use information from an onboard mapping pipeline which is continuously updated as the robot moves. 
We therefore require a fast update rate for our planner to keep up with map updates.
Building a planning graph incrementally by maintaining the graph between planning queries can speed up planning in static environments where most map updates will not invalidate the planning graph.
However, this approach necessitates graph maintenance between map updates, which adds heuristics and algorithmic complexity to identify updated graph elements.
The extreme challenges posed by the \acf{SubT}, like smoke, and dynamic obstacles, necessitate a more robust solution.

\subsection{Contributions}

In this work, we therefore introduce ArtPlanner, a navigation planner for legged robots which uses geometric reachability checking to find valid poses and a learned motion cost to find optimal paths which are safe and practically feasible.
\edited{This combination is enabled by a novel graph construction method:}
It creates a new planning graph every time the map is updated.
To keep computation time low, we lazily sample candidate pose vertices until we either reach a limit of number of poses, number of edges in the graph, or sampling time.
Because we plan on a robo-centric map of limited size, this allows us to densely sample the planning map while providing a limit on the sampling duration.
We then validate all graph edges at once by applying a locomotion risk threshold, leveraging massively parallel execution of a cost prediction network on GPU.
This allows for consistent and fast planning times.
The resulting paths are collision-free and can be followed safely by the used locomotion controller.

When robotic algorithms are deployed on hardware they have to deal with the imperfections of the real world and interact with other, possibly flawed parts of a robot software and hardware stack.
It is therefore beneficial to apply certain heuristics and make tweaks to part of the stack.
These ``small tricks'' are often glossed over but can be essential to making a system robust in practice.
For us, this was the case when computing the height map, which has the biggest impact on the outcome of planning decisions.
We detail three important components of our height map processing pipeline and show why each of them played an important role during the \acf{SubT} \textit{Finals}.

We extensively evaluated ArtPlanner during \ac{SubT} and present results gathered on four legged robots during the \textit{Finals}, for ArtPlanner and GBPlanner2, the planner used for exploration~\cite{kulkarni2022autonomous}.
We provide detailed analysis of the challenges faced by ArtPlanner, and how it managed to overcome these adverse conditions.
In addition, we compare our method to other state-of-the-art planners on the data gathered during the \textit{Finals} and show why other methods would not have been robust enough for \ac{SubT}.

Finally, we will open-source the code of our method upon acceptance of this work.

\subsection{Related Work}

Navigation planning for mobile robots is a vast field of research with a manifold of different approaches.

Most navigation approaches for mobile robots use a geometric environment representation as their basis for planning~\cite{wermelinger2016navigation,krusi2017driving,chavez2017image,oleynikova2017voxblox}.
They use various different terrain representations for planning, most commonly 2.5D height maps~\cite{wermelinger2016navigation,chavez2017image}, point clouds~\cite{krusi2017driving} or \ac{TSDF}s~\cite{oleynikova2017voxblox}.
Because planning in full 3D representations is currently computationally prohibited~\cite{tonneau2018efficient}, we chose to work with 2.5D height maps as the environment representation.
Most planning approaches compute a single geometric traversability value per terrain patch~\cite{wermelinger2016navigation,krusi2017driving} as measure for how easily the terrain can be traversed, irrespective of robot orientation.
Thereby, they neglect the much higher mobility which legged robots provide due to their ability to step over obstacles.
\edited{An overview of different traversability analysis approaches can be found in a recent survey article~\cite{borges2022survey}}.
While we have argued in previous work~\cite{wellhausen2019selfsupervised,wellhausen2020safe} that purely geometric approaches are not sufficient for navigation in natural outdoor environments, approaches relying on semantic information exhibit the same issues as traditional geometric approaches.
They, either implicitly through semantic segmentation of the environment~\cite{rothrock2016spoc,bradley2015scene,otsu2016autonomous,valada2017adapnet}, or explicitly~\cite{kim2006traversability,barnes2017find,hirose2018gonet} predict a traversability label.
However, we can instead reinterpret traversability labels as foothold feasibility labels and use them to enhance geometric planning.
While full kino-dynamic planning over long horizons would be the most general and accurate planning method, applying these methods in real-time is not tractable on current computational hardware~\cite{tonneau2018efficient,fernbach2017kinodynamic,winkler2018gait}.
Dynamic planning using a reduced robot model has recently shown promising results~\cite{norby2020fast} but has not been evaluated in deployment scenarios.
Other work on navigation planning specifically for legged robots either only considers cases of obstacle avoidance on flat terrain~\cite{zhao2018obstacle,harper2019energy} or does additional contact planning, which pushes computational complexity past the real-time mark~\cite{belter2019employing,lin2017humanoid,reid2020sampling}.
Approaches which learn traversability~\cite{chavez2017image} or motion cost~\cite{guzzi2020path,yang2021real} are powerful, but are either too slow due to the sequential querying of neural networks during sampling-based planning~\cite{guzzi2020path} or struggle in tight spaces where precise motion checking is necessary~\cite{yang2021real}.
In previous work~\cite{wellhausen2021rough}, we have used reachability planning with a learned foothold score to achieve real-time performance for legged navigation planning.
While resulting paths were generally feasible, the employed shortest path cost did not sufficiently account for locomotion risk on challenging terrain or close to obstacles.
For further related work of other navigation planners used during \ac{SubT}, please refer to \Cref{sec:related_work_subt}.

ArtPlanner uses a reachability-based robot representation~\cite{tonneau2015reachability} and learned foothold scores~\cite{wellhausen2021rough} with batched motion cost computation~\cite{yang2021real}.
This is the first work which combines geometric collision checking and learned motion costs in a navigation planner for legged robots.

\subsection{DARPA Subterranean Challenge}

The \acf{SubT} was a robotics challenge initiated by the \ac{DARPA} in 2018.
Its goal was to expedite development of robotic systems to rapidly map, navigate, and search complex underground environments such as human-made tunnel systems, urban underground and natural cave networks.
Three \textit{Circuit} events were held focusing on the aforementioned underground environments to qualify for the \textit{Finals} event in Louisville, KY which combined all environments into a single, purpose-built course.
Eight qualified teams, six of which were \ac{DARPA} funded, competed on September 21-23, 2021 to explore the unknown course to find, locate and identify a number of artifacts.
For every artifact which was identified correctly and localized accurately, a point was awarded.
Two preliminary rounds of 30 minutes were followed by the all-deciding 60 minute \textit{Prize Run}.
Only a single human supervisor was allowed to remotely interact with the robots once they entered the competition area and communications were severely limited, requiring a high level of autonomy from the robots.
The composition of the robot teams was not prescribed by \ac{DARPA}, and approaches changed over the four years of the competition.
In the end, all funded teams brought at least one legged robot to the \textit{Finals}.
Our team CERBERUS~\cite{tranzatto2022cerberus}, which won the competition, focused their ground robot efforts exclusively on legged robots from the beginning and brought four ANYmal-C~\cite{hutter2016anymal} quadrupeds to the \textit{Finals}.

\subsubsection{Navigation Planners in the SubT Challenge}
\label{sec:related_work_subt}

The approaches for ground robot navigation used during \ac{SubT} are diverse.
Team CoStar used a 2D multi-layer risk map to assess the terrain and planned the robot base path using a risk-aware kinodynamic MPC planner~\cite{fan2021step,thakker2021offroad}.
Team CSIRO Data61 used heuristic height map features to assess terrain traversability.
They also used the concept of virtual surfaces (see \Cref{sec:method_virtual_surfaces}) to compute a lower bound for terrain inclination which was used to avoid negative obstacles~\cite{hines2021virtual}.
Additionally, they used a deep reinforcement learning policy to control their tracked robots through narrow gaps~\cite{tidd2021narrow_gap}.
Team Explorer developed a kinodynamic local planner~\cite{cao2021tare} as well as a viewpoint-based planner using a polygonal representation of the environment~\cite{fan2021far}. 
However, both works do not explicitly state how the traversable regions and obstacles were computed.
Team CTU-CRAS-NORLAB computed traversable regions in a height map based on the neighboring cell height difference and planned using the A* algorithm~\cite{bayer2019autonomous,bayer2020modelling}.
Team MARBLE used an Octree representation of the environment and computed traversability based on the surface normal of extracted ground voxels~\cite{michael2021multi_agent}.
They employed a graph-based global planner combined with a reactive local controller.
All other funded teams besides team CERBERUS used a team of heterogeneous ground robots and therefore did not explicitly consider capabilities of legged robots.

\subsection{ANYmal Hardware}

\begin{figure}[t]
  \begin{subfigure}[t]{0.23\linewidth}
    \centering
    \includegraphics[width=\linewidth]{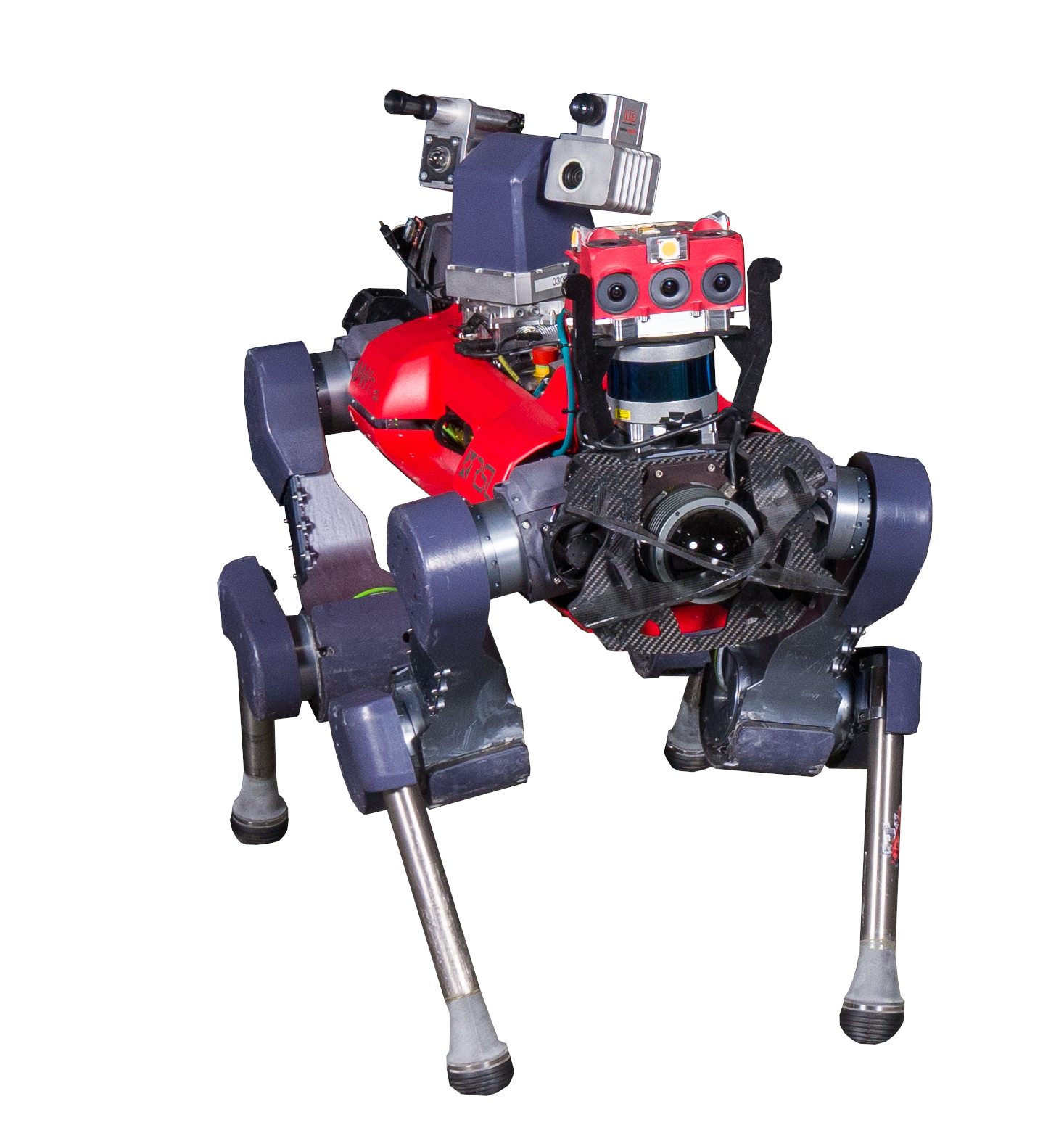}
    \caption{Explorer specification}
  \end{subfigure}
  \hfill
  \begin{subfigure}[t]{0.23\linewidth}
    \centering
    \includegraphics[width=\linewidth]{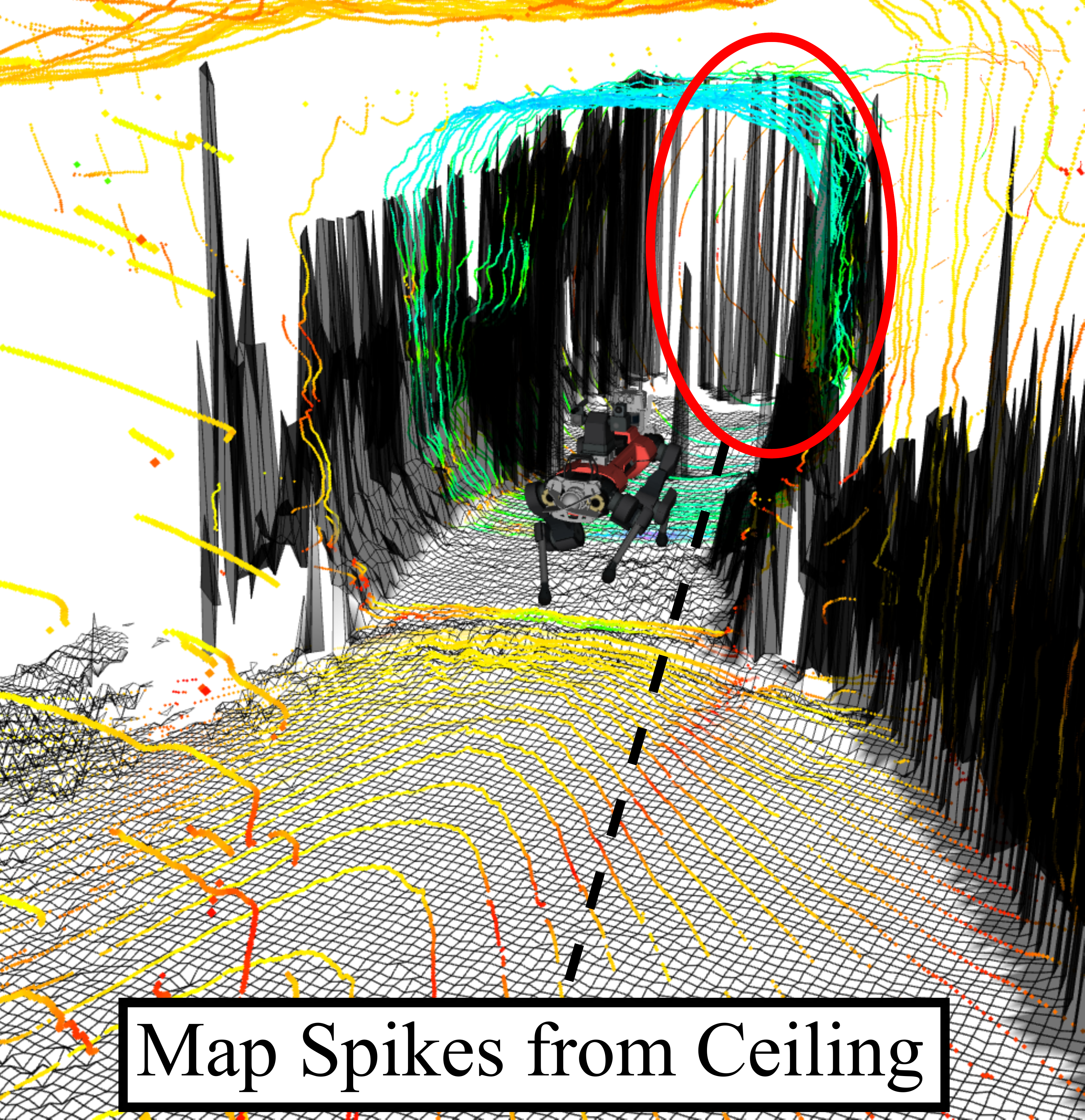}
    \caption{Explorer sensor data and height map}
  \end{subfigure}
  \hfill
  \begin{subfigure}[t]{0.23\linewidth}
    \centering
    \includegraphics[width=\linewidth]{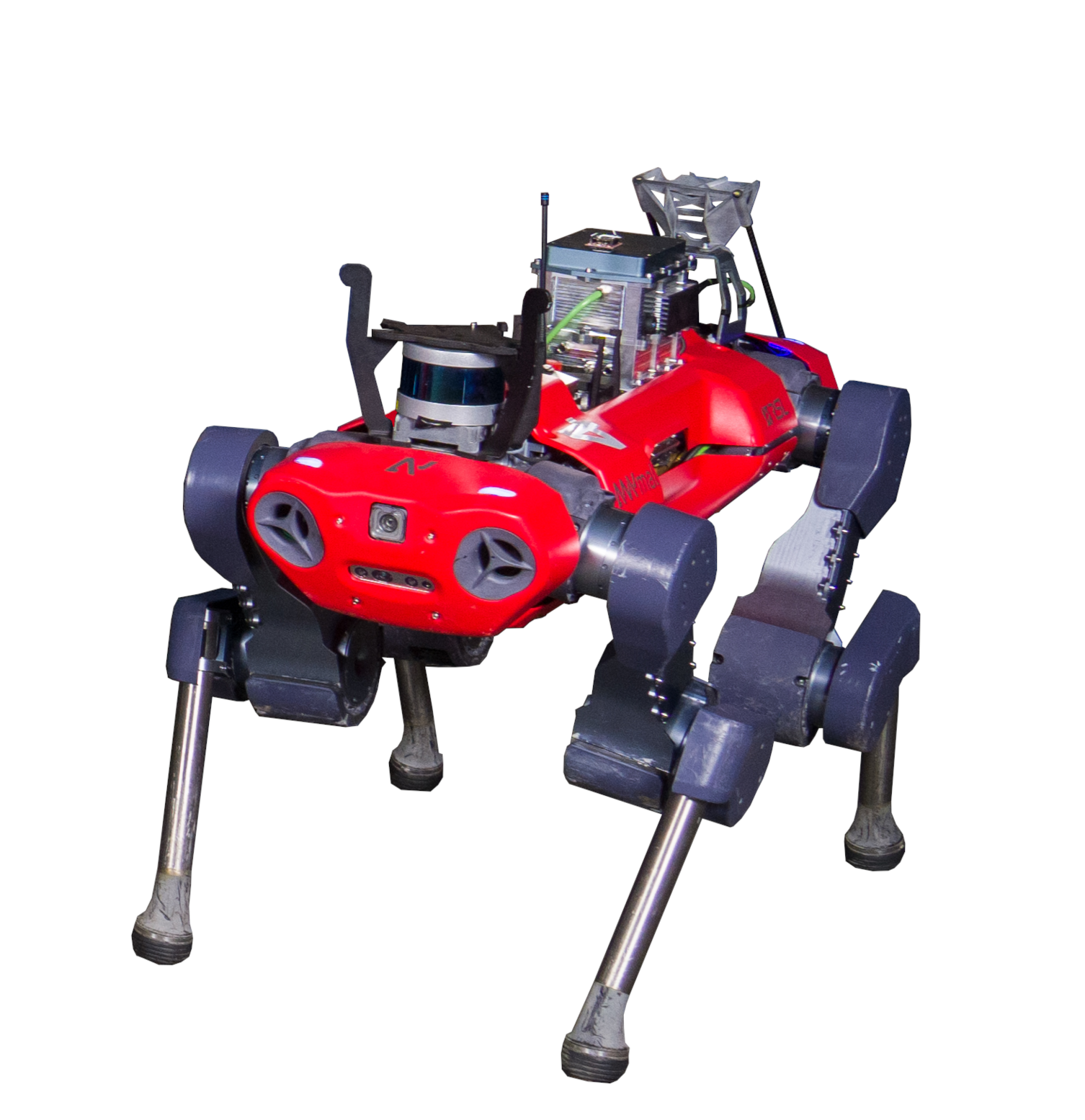}
    \caption{Carrier specification}
  \end{subfigure}
  \hfill
  \begin{subfigure}[t]{0.23\linewidth}
    \centering
    \includegraphics[width=\linewidth]{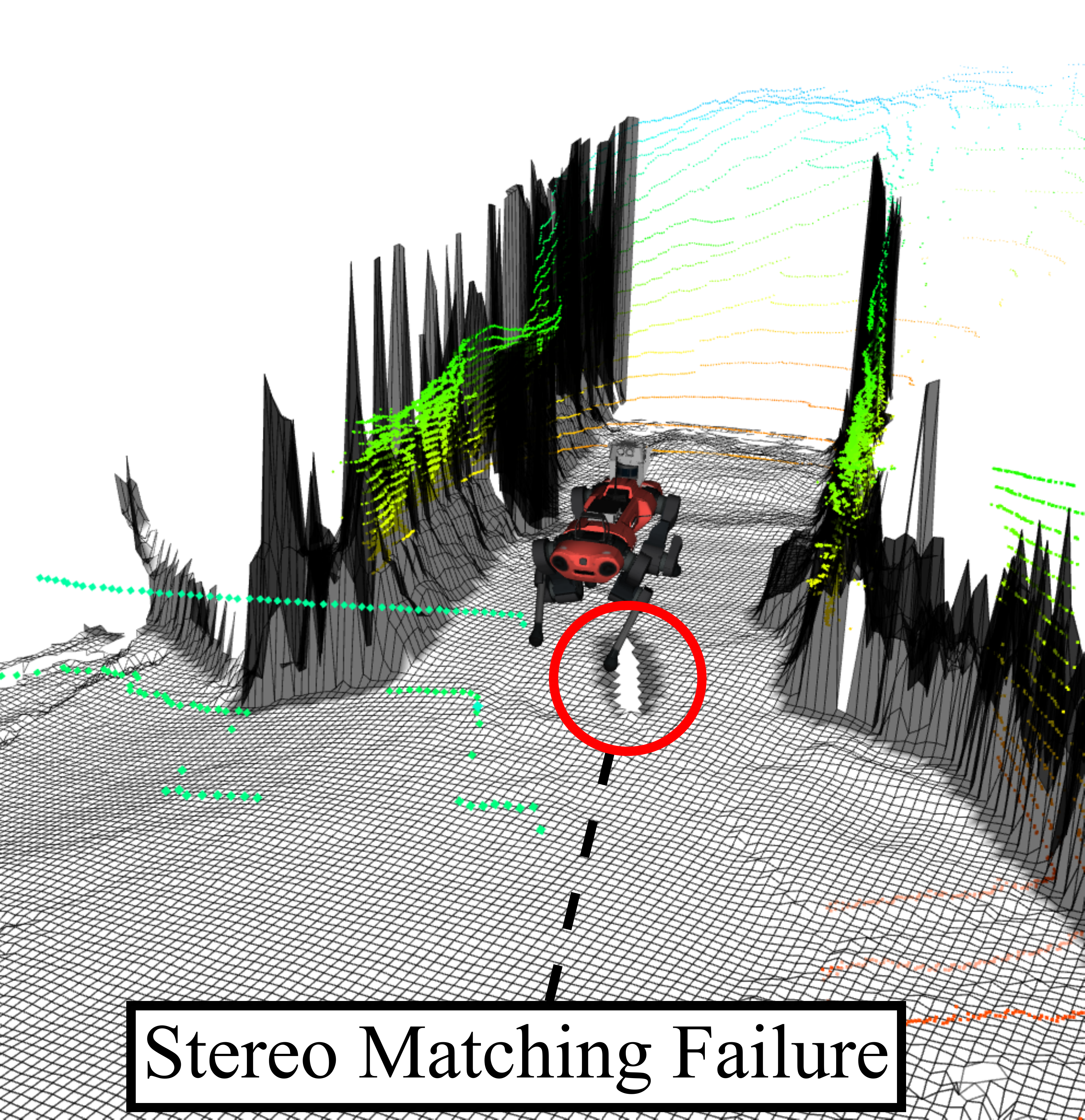}
    \caption{Carrier sensor data and height map (depth camera data not shown)}
  \end{subfigure}
  \caption{Modified ANYmal-C robots used during the competition.}
  \label{fig:anymal_hardware}
\end{figure}

For the \textit{Finals}, team CERBERUS customized four ANYbotics ANYmal-C~\cite{hutter2016anymal} robots with two different specifications.
All robots carried a Velodyne VLP-16 puck lidar for localization and mapping.
For compute they had two PCs with an Intel i7-8850H CPU as well as an Nvidia Jetson Xavier for GPU processing, connected via ethernet.
An \textit{explorer} specification robot is shown in \Cref{fig:anymal_hardware}(a).
They were equipped with two Robosense BPearl dome lidars placed at the front and the rear of the robots and a pan-tilt head on their back which housed various sensors and a lighting system for artifact detection.
\Cref{fig:anymal_hardware}(c) pictures a \textit{carrier} robot. 
They retained the four Realsense D435 depth cameras placed on each side of the base-spec ANYmal and were equipped with a mechanism for dropping WiFi mesh nodes.

In the context of this work, the most essential distinction between the two is the sensors used for height mapping of the environment.
The \textit{explorer} robot used its two dome lidars for height mapping, which return highly accurate and reliable readings.
However, its scan pattern is sparse and requires sweeping the ground through robot motion to obtain a dense height map.
Furthermore, its 180 degree \ac{FoV} means that the ceiling is also frequently observed, which can lead to issues with 2.5D height mapping (see \Cref{sec:results_map_spikes}).
The \textit{carrier} robot used its four depth cameras for height mapping which combined provide 360 degree coverage around the robot.
Unfortunately, their somewhat small stereo baseline meant depth accuracy dropped below a usable range around 2 meters from the robot.
Besides limited range, their active stereo matching technology can also lead to outliers on reflective and low-texture surfaces as well as missing data in environments with high lighting contrast.
On the \textit{carrier}, we therefore also used the Velodyne VLP-16 puck lidar to get reliable height readings farther from the robot.
Examples for the sensor data and the resulting height map for both specifications at the entrance to the Cave section of the \textit{Finals} course are shown in \Cref{fig:anymal_hardware}(b)+(d).
We used the same navigation planner with a single set of parameters for both robot specifications such that our method had to deal with their different height map characteristics.

\subsection{Navigation Stack}
\label{sec:intro_navigation_stack}

\begin{figure}[t]
  \includegraphics[width=\linewidth]{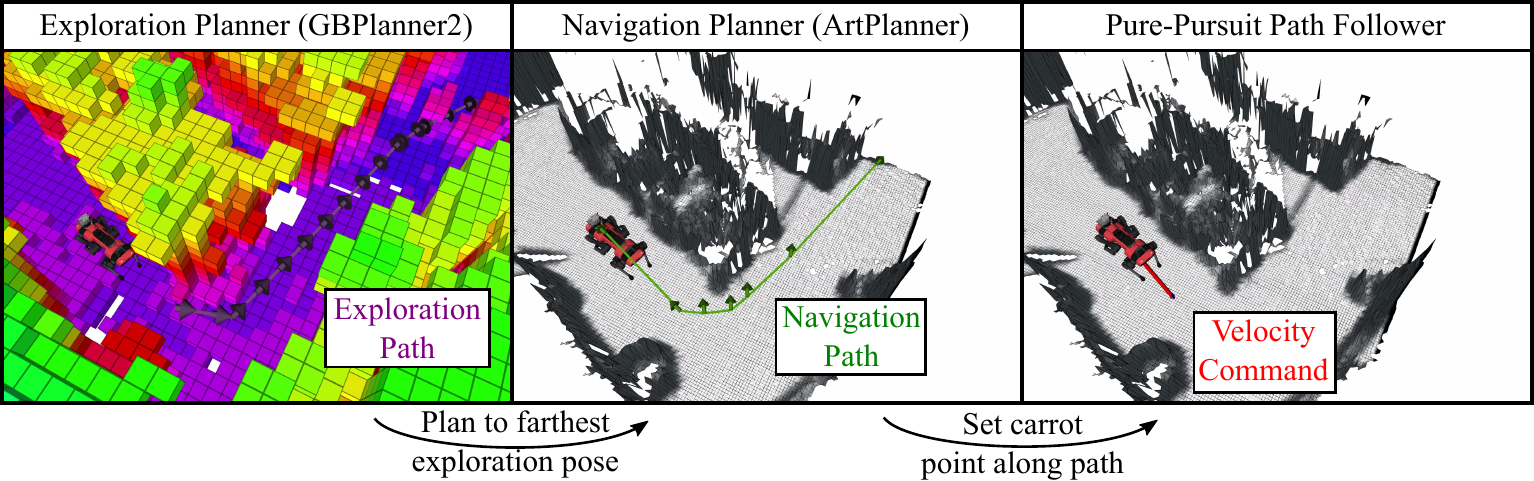}
  \caption{Team CERBERUS' navigation stack. An exploration planner~\cite{kulkarni2022autonomous} plans on a coarse volumetric map to maximize information gain. ArtPlanner then plans to the farthest reachable exploration pose on a more fine-grained robo-centric height map. Navigation paths are tracked using a pure-pursuit path follower.}
  \label{fig:navigation_stack}
\end{figure}

In the context of \ac{SubT}, ArtPlanner was embedded into a larger navigation stack, shown in \Cref{fig:navigation_stack}, to provide capabilities for autonomous exploration, and to actually follow computed paths.
These components were connected by a behavior tree to provide robustness and enable direct goal input to the navigation planner from the operator.
Details of this are out of scope of this work and can be obtained from the overview article of team CERBERUS~\cite{cerberus}.

Our graph-based exploration planner~\cite{dang2019graph,kulkarni2022autonomous} (GBPlanner2) plans to maximize information gain along the robot path.
It operates on a global 20\si{\centi\meter} voxel-size volumetric map~\cite{oleynikova2017voxblox} which is too sparse to capture the terrain in sufficient detail for legged robot navigation.
Knowing that we had ArtPlanner to provide safe navigation, we also tuned the parameters of GBPlanner2 to be optimistic and favor exploration gain over safety considerations. 
Specifically, this meant we used a small collision volume for validity checking without any safety margin to fit through narrow openings.
Additionally, similar to the concept of virtual surfaces~\cite{hines2021virtual}, GBPlanner2 is allowed to output so called ``hanging vertices'', where vertices in free space are not supported by any ground surface underneath, but only unobserved space (see \Cref{sec:results_safety_threshold}).
We therefore need our navigation planner to refine the exploration path in cases where its low-resolution map causes suboptimal or risky paths and to stop the robot if the path is completely infeasible.
For this, we plan on a local height map which is centered at the current robot positions and moves with it.

When ArtPlanner receives a new exploration path, it iterates through the path poses in reverse, starting with the farthest one, and tries to plan to each pose.
This is done to maximize the planning distance, such that ArtPlanner can optimize the path and circumvent any obstacles which might have been missed by the exploration planner.
We continuously repeat this at a target planning rate of 0.5\si{\hertz}.
When we successfully plan to an exploration path pose, all path poses which precede it are considered reached and are not used for future planning iterations.
Planning of the GBPlanner2 is triggered in two cases: 
Either if the exploration path is infeasible and therefore contains no pose we can plan to.
Or, if the last pose in the exploration path has been reached.

The resulting navigation path is tracked using a pure-pursuit PID path follower.
It considers both the horizontal distance as well as the angular distance when setting the carrot point along the path.
This is done to ensure accurate tracking in tight spaces and speedy locomotion when moving forward.
Since we used a single robust and perceptive locomotion controller~\cite{miki2022learning} we did not have to consider any gait or locomotion controller switching.


\section{Method}
\label{sec:method}

ArtPlanner uses reachability-based pose validity checking with learned foothold scores and computes path costs using a learned motion cost module.
Because a high quality map of the environment is crucial for good path planning, we perform additional processing on the input height map to increase its quality and improve safety.

\subsection{Reachability Planning}
\label{sec:method_base_planner}

\begin{figure}[t]
  \begin{subfigure}[t]{0.32\linewidth}
    \centering
    \includegraphics[width=\linewidth]{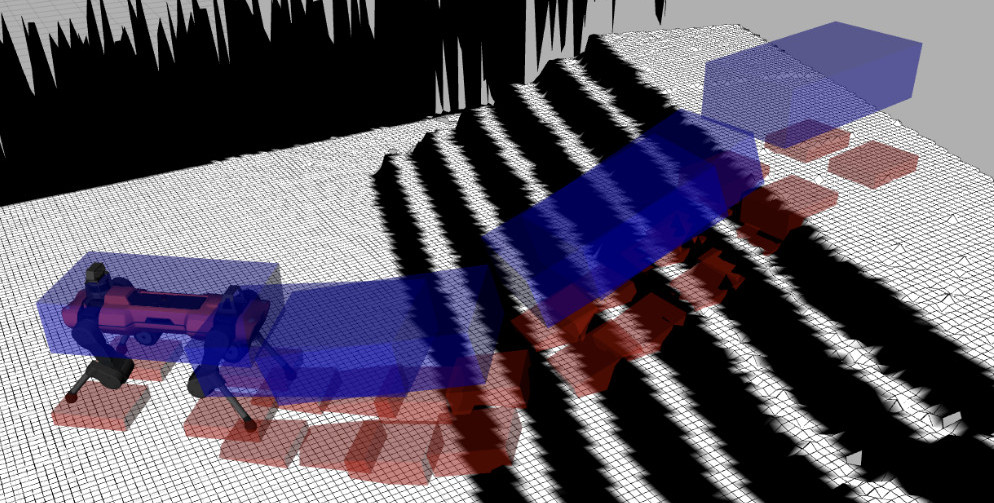}
    \caption{Reachability Abstraction}
  \end{subfigure}
  \hfill
  \begin{subfigure}[t]{0.32\linewidth}
    \centering
    \includegraphics[width=\linewidth]{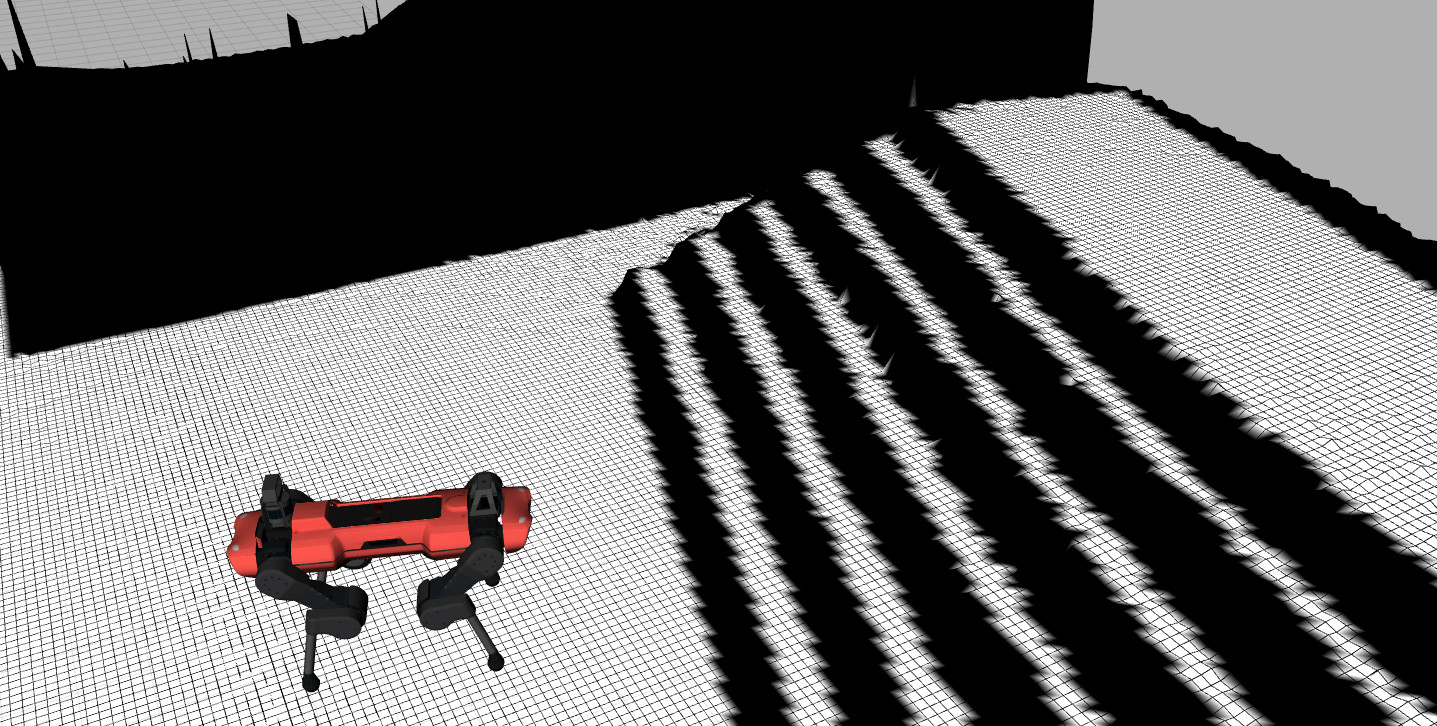}
    \caption{Torso Collision World}
  \end{subfigure}
  \hfill
  \begin{subfigure}[t]{0.32\linewidth}
    \centering
    \includegraphics[width=\linewidth]{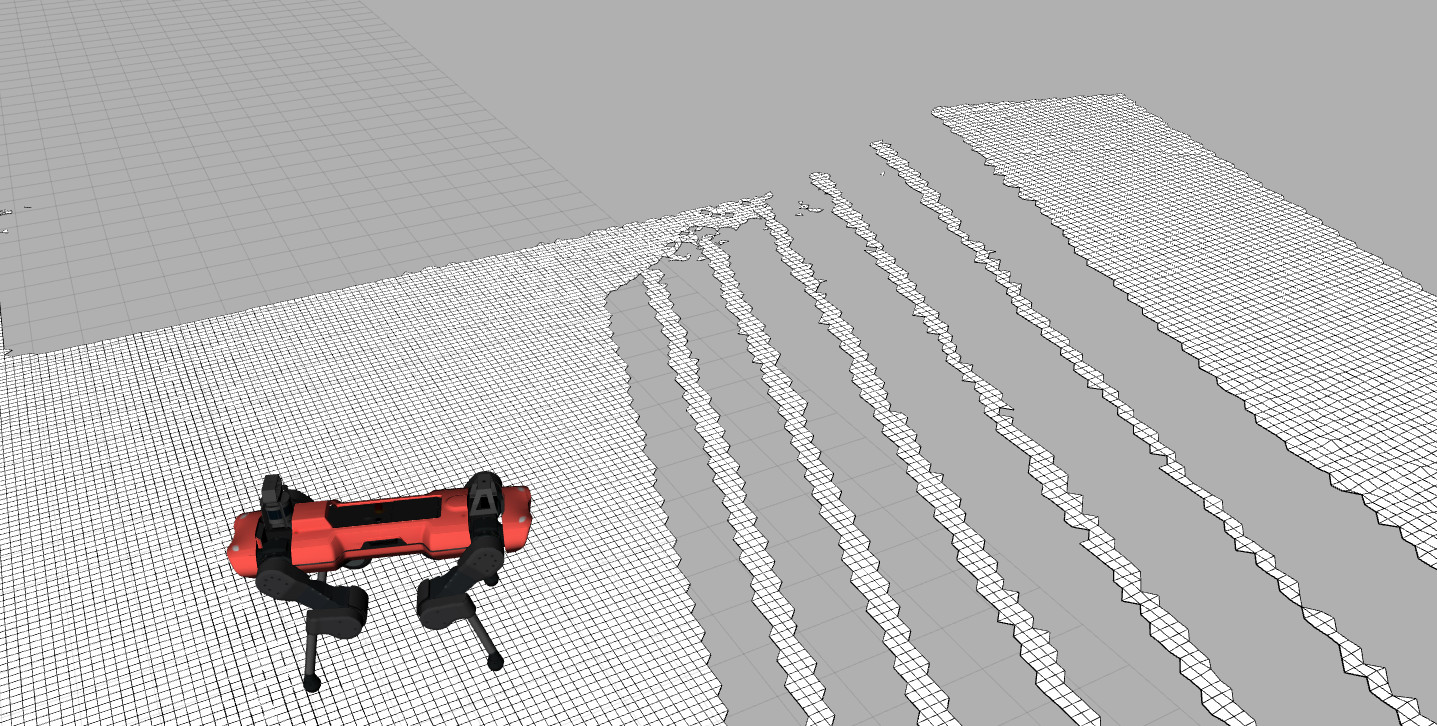}
    \caption{Reachability Collision World}
  \end{subfigure}
  \caption{We use a reachability abstraction of the robot (a). Blue boxes need to be contact free in the torso collision world while red boxes need to be in contact with the reachability collision world. The torso collision world (b) considers all geometry perceived by the mapping algorithm. Regions considered unsuitable for footholds are colored in black. This geometry is removed (c) when checking reachability volumes for collision.}
  \label{fig:reachability_planning}
\end{figure}

To check for validity of sampled robot poses, we use a reachability-based approach~\cite{tonneau2015reachability}.
It is based on the notion that the ground support surface needs to be reachable for the robot's legs, while the torso remains collision-free.
To check for this condition, we decompose the robot body into volumes representing torso collisions and leg reachability, as shown in \Cref{fig:reachability_planning}(a). 

In order to account for geometry which should not be used as support surface (i.e. walls), we trained a \ac{CNN} to predict a foothold score based on height map information and use this to constrain the regions considered for valid footholds~\cite{wellhausen2021rough}.
Because the \ac{CNN} is very small, with few parameters, we can train it with only 20 hand-labelled samples.
The foothold score is indicated with black and white color in \Cref{fig:reachability_planning}.
Geometry which has a low foothold score is disregarded for collision checking of the limb reachability volumes but is still considered for the torso, visualized in \Cref{fig:reachability_planning}(b)+(c).
A valid pose is therefore a pose where all reachability volumes are in contact with valid geometry, while the torso volume is not in collision with any geometry.

Inspired by the sampling-based LazyPRM*~\cite{hauser2015lazy} algorithm, ArtPlanner only checks the sampled pose for validity when adding a new node to the graph, but does not immediately check the validity of newly added edges.
Edge checking is performed using batched motion cost computation when a path is queried, as discussed in \Cref{sec:method_motion_cost}.
We use a custom sampling scheme which uses a 2D pose sample augmented to a 3D pose using map information and biases sampling towards regions with low node density~\cite{wellhausen2021rough}.

\subsection{Learned Motion Cost}
\label{sec:method_motion_cost}

A simple shortest-path cost can lead to risky paths which require the robot to step close to obstacles and edges.
Due to the difficulty of analytically determining locomotion cost and risk, other recent work has proposed to use a learned neural network to compute cost~\cite{guzzi2020path}.
Building upon these works, we use a neural network which computes the energy and time required and failure risk associated with moving from a query location in the height map to a given relative 2D pose~\cite{yang2021real}.

The inputs to the network are a patch of the height map centered around the robot, the current yaw orientation of the robot, as well as the 2D goal pose relative to the robot.
The network outputs the time $c_t$ and energy $c_e$ required to move to the target pose as well as the motion risk $c_r$, which is the probability that this transition fails.
The network is trained using data generated in simulation, where the robot is spawned in a random location on a randomly generated height map and a goal pose in a certain range around the robot is randomly selected.
We then naively give a directional command to the robot to move towards this goal pose while the robot experiences external disturbances.
This is repeated a number of times to compute the failure probability.
Time and energy values are averaged over all attempts which successfully reached the goal.

The risk value is used to determine validity of planning graph edges, while the graph edge cost is computed as a weighted sum of time $c_t$, energy $c_e$ and risk $c_r$:

\begin{equation}
\label{eq:cost}
    c = w_t\cdot c_t + w_e\cdot c_e + w_r\cdot c_r
\end{equation}

$c_t$ and $c_e$ are normalized with their respective maximal value in the training dataset, whereas $c_r$ expresses the probability of failure.
Battery runtime of the ANYmal robots was not of concern and energy consumption $c_e$ showed to be highly correlated with time $c_t$, so we chose to disregard this cost term.
Since any navigation error would be potentially mission-ending during the \ac{SubT} \textit{Finals}, we put a high weight on risk.
Consequently, our chosen cost weights were $w_t:=1$, $w_e:=0$ and $w_r:=5$.

\subsection{Graph Construction}
\label{sec:method_graph_construction}

\begin{figure*}[t]
  \includegraphics[width=\linewidth]{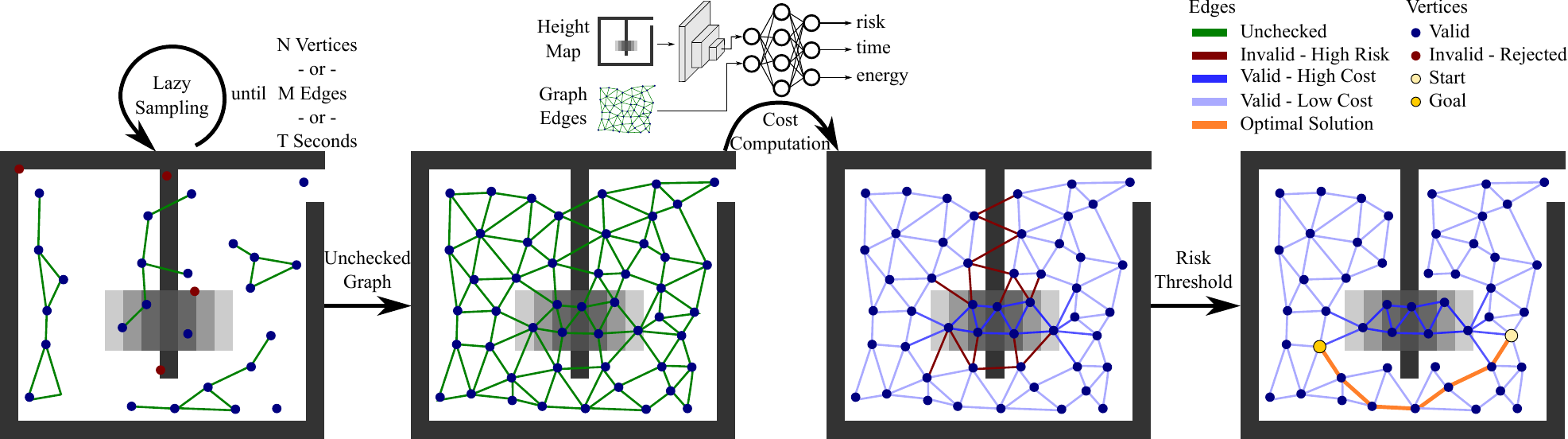}
  \caption{Our planner lazily samples graph poses, only checking state but not edge validity. Once a sampling limit is reached, the motion cost for every graph edge is computed in a batched fashion. The graph is pruned by applying a motion risk threshold and the optimal path can be found using an A* search.}
  \label{fig:planner_overview}
\end{figure*}

We construct our planning graph using lazy sampling, where only the added vertices but not edges are immediately checked for validity.
When querying a solution in a regular LazyPRM* graph, an A* search is performed and all edges which are part of the optimal solution are checked for validity.
Any invalid edges are removed from the graph and the process is repeated until either all edges returned by the A* search are valid (success), or the start and goal vertices are no longer connected (failure).
This makes the path query time highly non-deterministic since the number of A* search iterations strongly depends on the complexity of the environment which determines the number of invalid edges.
As discussed in our previous work~\cite{wellhausen2021rough}, this can lead to excessive planning times which make real-time applications impractical or impossible.
While our previously suggested graph expansion method~\cite{wellhausen2021rough} can compensate for this issue in most cases, during extensive real-world testing in preparation for the \ac{SubT} challenge we still encountered some instances of long planning times in  difficult environments where the number of invalid edges in the graph is exceedingly large.
\edited{In addition, when used with a learned motion cost, classical sampling-based planning requires inefficient, sequential querying of the motion cost network.
This drastically increases planning times~\cite{guzzi2020path}.}

To achieve consistent and low planning times, we therefore leverage batched edge cost computation using our motion cost network, detailed in \Cref{fig:planner_overview}.
Every time a new planning query is received and the map has updated, we build a new planning graph by lazily sampling until the graph exceeds either $N$ valid vertices, $M$ unchecked edges or we have sampled for more than $T$ seconds.
The number of graph edges determines the batch size for the motion cost query, which in turn determines the inference time of the motion cost network.
Therefore, the upper bound of unchecked edges $M$ limits the maximal computation time used by the cost query.
The other two termination criteria prevent unnecessarily long and dense sampling in certain environments where either the traversable area of the map is very small ($T$ limit) or the geometry allows every node to only have a small number of neighbors ($N$ limit), like in narrow corridors.

A batched query of the cost network is then performed for every edge in the graph at once, which can be executed efficiently on GPU.
Edges which exceed a risk threshold $R$ are removed and valid edges are assigned a cost, according to \Cref{eq:cost}.

We now have a fully validated graph with assigned edge costs and consequently, a single A* search will return the the optimal path between two query nodes.

\subsection{Height Map Processing}
\label{sec:method_map_processing}

The height map provided to the planner is its only source of information about the environment.
Obtaining a high quality height map is therefore paramount for safe planning.
To achieve this, we perform additional processing steps to improve map quality and increase planning safety.

\begin{figure}[t]
  \includegraphics[width=\linewidth]{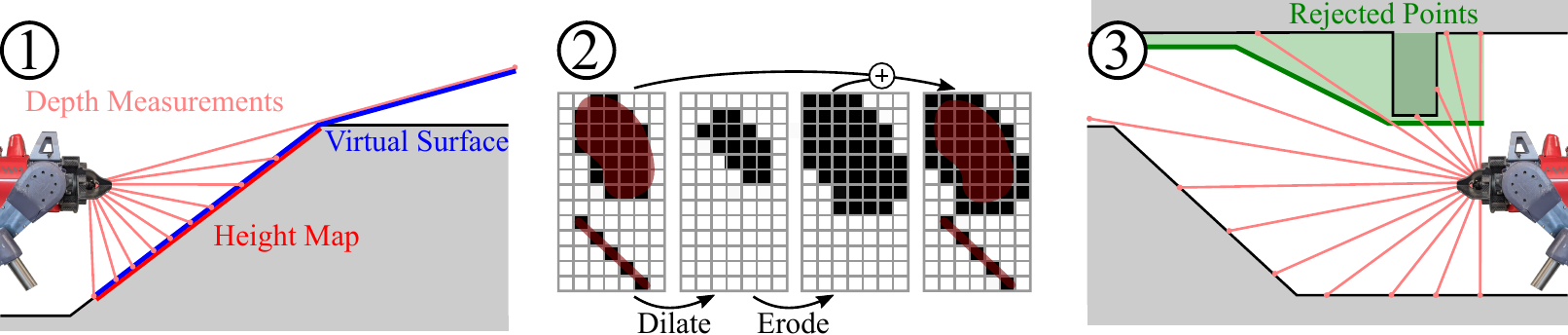}
  \caption{\mynumber{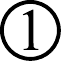}~We use depth measurement rays to infer virtual surfaces above the robot in unobserved map space. \mynumber{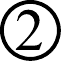}~By first dilating steppable map terrain and then eroding it, we compute a safety threshold around dangerous terrain (red). \mynumber{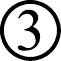}~We reject ceiling points when computing the height map using a height threshold which changes with distance from the robot, to handle both low openings and steep inclines.}
  \label{fig:map_processing}
\end{figure}

\subsubsection{Virtual Surfaces}
\label{sec:method_virtual_surfaces}

The depth sensors on our legged robots are mounted about half a meter above the ground.
Especially on rough terrain and inclines, this can often lead to large occlusions in the height map.
To combat this issue, we compute virtual surfaces~\cite{hines2021virtual} as the upper bound of a height value in unobserved cells, shown in \Cref{fig:map_processing}~\mynumber{1}.

We cast a ray from the sensor origin to every point in the observation point cloud.
For every cell this ray passes through we know that its height value cannot be larger than the height at which the ray passes through the cell.
This way, we can fill unknown map regions using sensor information, instead of simply filling them using an image inpainting algorithm or considering them untraversable.

However, the low sensor placement means that when we approach a negative obstacle, like a cliff, its virtual surface will appear to have a very small inclination, until we are very close to it.
This introduces additional risk, if the planner and path follower are not able to stop the robot in time, once the severity of the drop appears in the virtual surface.
We therefore decided to only use virtual surfaces for planning if they are above sensor height.
This way, we can benefit from them if we are walking up inclines (see \Cref{sec:results_virtual_surfaces}), but also stay safe in the presence of negative obstacles (see \Cref{sec:results_safety_threshold}).

\subsubsection{Safety Threshold}

We rely on the learned motion cost network to compute the path risk to avoid walking too close to dangerous obstacles.
While this works well if the environment is mapped well, negative obstacles often do not appear in the height map, as discussed above in \Cref{sec:method_virtual_surfaces}.
We can therefore not rely on the motion cost network to keep a safe distance from negative obstacles.

To combat this issue, we apply a safety margin to disallow stepping too close to edges, illustrated in \Cref{fig:map_processing}~\mynumber{2}.
It is implemented using image erosion on the foothold score layer of the height map, which reduces the steppable map region by a safety margin.
This has the additional benefit of also removing small isolated steppable patches from the map, which could only be overcome by solving a stepping-stone problem.
To avoid unnecessarily inflating small obstacles like rails, which the robot can easily step over, we do not inflate unsteppable regions below a certain size. 
In practice this is done by performing an image dilation of smaller radius, before doing the erosion.

This safety threshold was crucial on the Subway Station of the \textit{Finals} circuit, as shown in \Cref{sec:results_safety_threshold}.

\subsubsection{Ceiling Point Filter}
\label{sec:method_ceiling_point_filter}

We use a 2.5D height map representation for planning.
While this is sufficient for ground robot navigation in most environments, it can be problematic in the tight underground spaces encountered during \ac{SubT}.
Low ceilings mean that they are frequently observed by the depth sensors, which causes spikes in the height map, as shown in \Cref{fig:anymal_hardware}(b).
However, we cannot simply discard all points above a fixed height, since this would either prevent us from planning up slopes or from passing underneath low overhangs.

We therefore use a rising height threshold to filter points~\cite{miki2022elevation}, shown in \Cref{fig:map_processing}~\mynumber{3}.
It filters points just above robot height close to the robot, and linearly increases the height threshold up to a maximum at larger distances.
This setup caused map spikes in parts of the course with low ceilings which slowed us down, but these crucially never stopped us from exploring. 
It allowed us to pass underneath very low overhangs, and to plan up slopes, even when encountered together, as detailed in \Cref{sec:results_map_spikes}.


\section{Experimental Results}
\label{sec:result}

ArtPlanner was deployed on all four ANYmal-C ground robots of team CERBERUS during all runs of the \ac{SubT} \textit{Finals}.
It used a height map of size \SI{8}{\meter}$\times$\SI{8}{\meter} with a \SI{4}{\centi\meter} resolution.
We only cover results related to the navigation planner presented in this work.
For further details on the general performance we refer to our overview article~\cite{cerberus}.

We deployed all four ground robots during the \textit{Prize Run}, which were directed by the supervisor to explore different areas of the course.
All ground robots successfully made it to the end of the competition and we did not observe a single \edited{path planning} or locomotion failure, which could have been provoked by bad path planning.
Our planner was active for 90 minutes between all robots which accounts for 88.94\% of all robot motion.
We gracefully navigated the narrow doorways and small rooms in the Urban section, passed through the Tunnel section with obscuring fog, and made it through the narrowest and roughest part of the Cave section.
\edited{The only case where ArtPlanner did not follow the exploration path over traversable terrain happened at the stairs leading to the subway station. 
This resulted from our operational decision to use older, well-tested motion cost network weights, which produced elevated risk levels on stairs. We preferred these weights over newer, untested weights, which perform well on stairs, since stairs were not a prominent feature in the Finals course.}
We only collided with the environment a single time, getting caught on a narrow pole due to path following delays, discussed in further detail in \Cref{sec:results_path_follower}.
The only other issue encountered by our planning method was slow progression due to artifacts in the height map, as outlined in \Cref{sec:results_map_spikes}.
\edited{Nevertheless, we safely explored large parts of the competition course, as shown in \Cref{fig:cost_maps}.}

\subsection{Comparison}

Directly following GBPlanner2's exploration path~\cite{dang2019graph,kulkarni2022autonomous} during the competition would almost certainly have lead to major issues.
It did not sufficiently account for traversability characteristics of the terrain, planning directly over high rails shown in \Cref{fig:cost_maps}~\mynumber{1}.
It even completely missed some smaller obstacles, like the traffic cones shown in \Cref{fig:cost_maps}~\mynumber{2}.
This could have possibly been avoided with more conservative tuning of the exploration planner, but this would have significantly impeded exploration.

To show that ArtPlanner was crucial to performing well in the challenging environment of the \ac{SubT} \textit{Finals}, we compare it to other navigation planning methods, using data collected during the Prize Run.
\Cref{fig:cost_maps} shows the motion cost and collision rate for all robots and all planners as a heat map overlayed over the top-view of the competition course.

\subsubsection{Other Planners}

Naturally, only ArtPlanner was running during the \textit{Finals}.
We recorded all paths planned during the \textit{Finals} to evaluate performance post-event.
To compare with other methods, we played back the state estimation, localization and height map data recorded during the \textit{Finals} and  input the exploration paths commanded during the \textit{Finals} to all planners.
This allows us to somewhat reproduce behavior of the first two components of the navigation stack, as shown in \Cref{fig:navigation_stack}, and therefore evaluate path quality.
Note, however, that this means planning will always start from the competition robot pose, which is a result of following our planner.
Therefore, planning always starts from a safe state and other intricacies which can lead to issues (see \Cref{sec:results_path_follower}) cannot be reproduced.
However, if planners show problematic behavior even in this simplified setup, they are even more likely to fail during real deployment.
To have a fair comparison to the other methods, we also re-ran ArtPlanner on \textit{Finals} data.

We evaluate the following methods:
\begin{itemize}
    \item \textit{ArtPlanner (Competition)}: Our method as run in the loop, on the robots, during the competition.
    \item \textit{GBPlanner2}: Exploration planner~\cite{kulkarni2022autonomous} as run during the competition.
    \item \textit{ArtPlanner (Playback)}: Our method using data from the competition played back.
    \item \textit{No Motion Cost}: A reachability-only planner~\cite{wellhausen2021rough}, which we fed the height map processed as proposed in this work.
    \item \textit{Motion Cost Planner}: Planner based purely on the motion cost network~\cite{yang2021real}. It first computes a raw path, by querying motions in a fixed graph pattern, with heuristics to determine robot orientation. This raw path is then optimized with gradient-based optimization using the motion cost network.
    \item \textit{Exploration Path w/ Cost Optimizer}: GBPlanner2~\cite{kulkarni2022autonomous} already outputs a path which is mostly generally feasible, but optimizes for information gain and plans on a coarser map. We therefore fed the exploration path directly into the motion cost optimizer~\cite{yang2021real} to obtain a refined navigation path.
\end{itemize}

\subsubsection{Methodology}
\label{sec:results_methodology}

\begin{figure}
  \includegraphics[width=\linewidth]{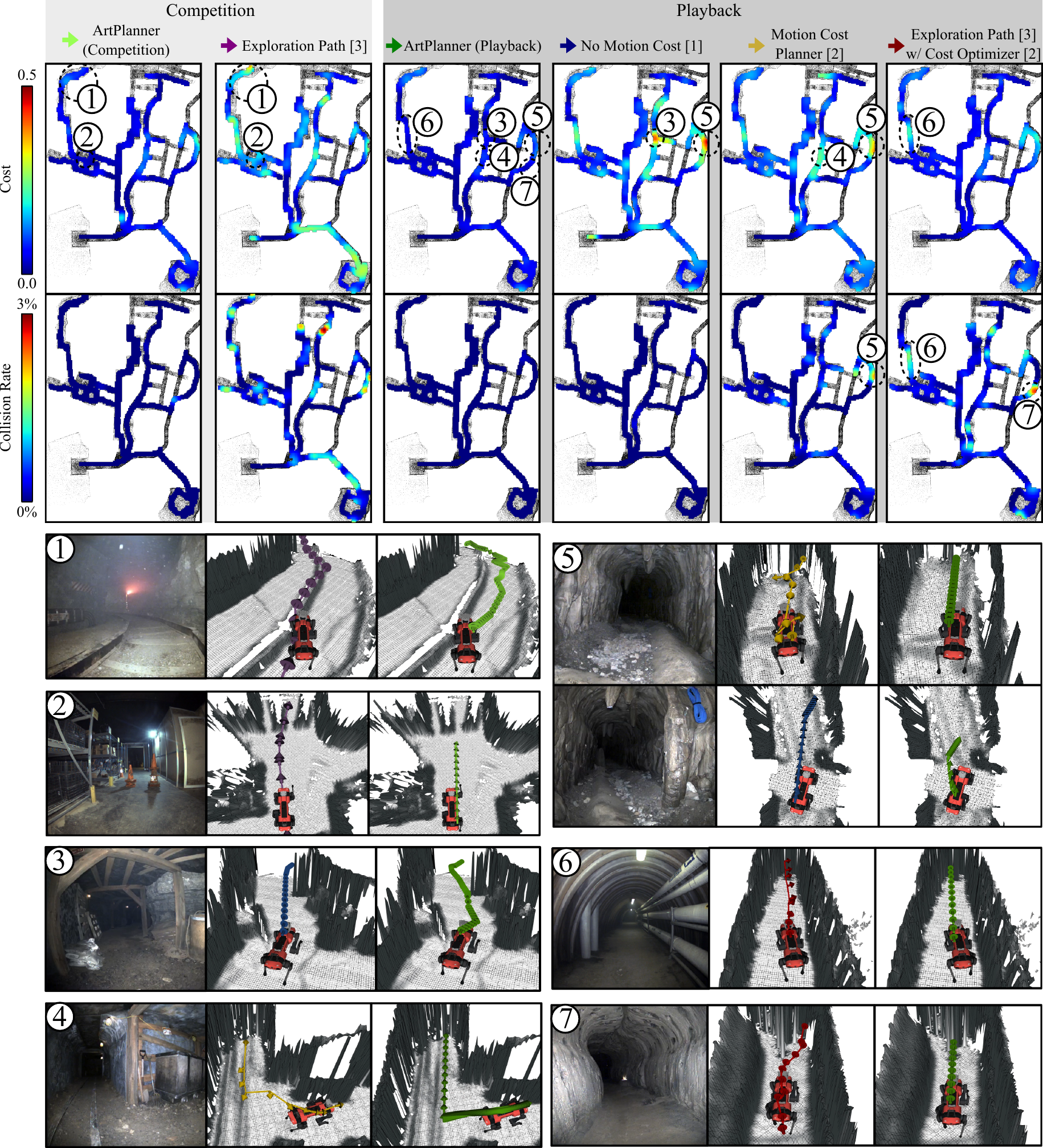}
  \caption{Comparison of path costs and path collision rate for different planners during the \ac{SubT} Finals Prize Run. The two left-most columns show data from the planners as run on the robots during the competition, while the other columns show real robot data played back. Data is smoothed with a Gaussian filter with standard deviation of 2\si{\meter} for better visibility.
  \mynumber{1} The \textit{Exploration Path} was highly risky whereas ArtPlanner avoids the high rail track. \mynumber{2} The \textit{Exploration Path} was infeasible as it ignored the obstructing traffic cones. \mynumber{3} With \textit{No Motion Cost} the path was very close to obstacles whereas ArtPlanner kept a safe distance. \mynumber{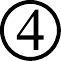} Due to a bad initial guess, the \textit{Motion Cost Planner} turned on the train tracks whereas ArtPlanner turned beforehand and walked straight on the tracks. \mynumber{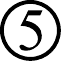} Due to a bad initial guess the \textit{Motion Cost Planner} performed an unnecessary turn, which caused the path cost optimizer to produce a colliding path in the narrow cave tunnel. \mynumber{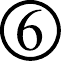} The initial exploration path was in collision with height map artifacts, leading to suboptimal behavior of the path optimizer. \mynumber{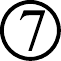} In trying to avoid map noise, the path optimizer pushed the path into the tunnel wall.\\~
  [1]~\cite{wellhausen2021rough}
  [2]~\cite{yang2021real}
  [3]~\cite{kulkarni2022autonomous}}
  \label{fig:cost_maps}
\end{figure}

We used the motion cost network to compute path costs for all path segments inside the height map.
The same network architecture and weights were used for evaluation and in the planners which use the cost network themselves.

Since the planners mentioned above have different characteristics, implementation details, and produce paths of different lengths, we took extra considerations to guarantee a fair comparison.
We check height map collisions only for the torso of the robot, analogous to the blue boxes shown in \Cref{fig:reachability_planning}(a).
We also reduced the size of these boxes by 10\si{\centi\meter} in all dimensions compared to the size used by the reachability planners.
This was done to account for the more accurate collision models used when training the motion cost network.
Finally, we disregarded the first pose of all output paths, since it corresponds to the current robot pose.
This means that it has to be valid and any detected collisions would be erroneous.
The reachability planners account for this by sampling a valid pose in a small region around the start, but the motion cost planners do not have that functionality.
When evaluating the \textit{Exploration Path}, we only consider path poses from the one closest to the current robot pose until the current goal pose of the ArtPlanner (see \Cref{sec:intro_navigation_stack}).
This prevents excessive collision rates caused by height map artifacts, which do not appear in the volumetric map used by GBPlanner2.
Note that all these concessions disadvantage ArtPlanner, and the comparison would have been even more in our favor without them.

When computing cost and collisions, we used the height map at the time when the paths are published, not the maps which they are planned on.
This is a more realistic evaluation which takes into account effects caused by different planning speeds of the methods.
We use the same risk cutoff of 0.5 for all methods and use the same cost function mentioned in \Cref{sec:method_motion_cost}.

All playback experiments were run on a Desktop computer with an Intel i7-8700K CPU, a Nvidia RTX 2080 GPU and 32 GB of RAM.

\subsubsection{Motion Cost}

To analyze the planner performance with respect to motion cost, we aggregate the cost of all individual path segments spatially.
We visualize this as a heat map, overlaid over a top-down view of the \textit{Finals} course map, shown in \Cref{fig:cost_maps}. 
\edited{We also show motion cost and motion risk statistics in \Cref{tab:collision_rate}}.

ArtPlanner consistently output paths with low motion costs in all regions of the map, as shown in \Cref{fig:cost_maps} \mynumber{1}-\mynumber{7}.
During the competition, ArtPlanner produced paths with slightly higher motion costs in very narrow parts of the course, most likely due to the lower compute budget available onboard the robot.
The \textit{Exploration Path w/ Cost Optimizer} generally also performed well in terms of cost and is able to reduce the high motion cost of the raw \textit{Exploration Path} in most cases.
There are no hot spots with high costs and it only produced elevated costs in two cases:
In the very narrow cave section and in one case where the exploration path intersected map artifacts, which lead to suboptimal optimization behavior (\Cref{fig:cost_maps} \mynumber{6}).
Interestingly, the full \textit{Motion Cost Planner} produced higher cost paths in more regions than the \textit{Exploration Path w/ Cost Optimizer}, even though it considers the motion cost at all planning stages. 
This is due to the orientation heuristics in the first planning stage, which can produce raw paths with motions the gradient-based optimizer cannot fix, like 360-degree turns.
This caused the path to rotate on challenging terrain (\Cref{fig:cost_maps} \mynumber{4}) and in narrow spaces (\Cref{fig:cost_maps} \mynumber{5}).
Unsurprisingly, the reachability planner with \textit{No Motion Cost} produced multiple dangerous, high cost hot spots, frequently planning very close to obstacles (\Cref{fig:cost_maps} \mynumber{3}+\mynumber{5}) and over rough terrain.

\subsubsection{Collision Rate}

\begin{table}
\centering
\begin{tabular}{ l | S[table-format=2.2] S[table-format=2.2] | S[table-format=2.2] S[table-format=2.2] | S[table-format=2.2] S[table-format=2.2]}
 & \multicolumn{2}{c|}{Collision Rate [\%]} & \multicolumn{2}{c|}{Motion Cost} & \multicolumn{2}{c}{Motion Risk} \\
  & \text{Any} & \text{Severe} & \text{Mean} & \text{95\%} & \text{Mean} & \text{95\%} \\ 
 \hline\hline
 \textit{ArtPlanner (Competition)} & 4.50 \cellcolor[rgb]{1,0.96,0.96} & 0.05 \cellcolor[rgb]{1,0.99,0.99} & 0.22 \cellcolor[rgb]{1,0.93,0.93} & 1.00 \cellcolor[rgb]{1,0.92,0.92} & 0.03 \cellcolor[rgb]{1,0.94,0.94} & 0.17 \cellcolor[rgb]{1,0.92,0.92} \\
 \textit{Exploration Path} & 10.93 \cellcolor[rgb]{1,0.89,0.89} & 2.44 \cellcolor[rgb]{1,0.76,0.76} & 0.77 \cellcolor[rgb]{1,0.74,0.74} & 4.34 \cellcolor[rgb]{1,0.64,0.64} & 0.12 \cellcolor[rgb]{1,0.76,0.76} & 0.74 \cellcolor[rgb]{1,0.63,0.63} \\ \hline
 \textit{ArtPlanner (Playback)} & 6.26 \cellcolor[rgb]{1,0.94,0.94}  & 0.34 \cellcolor[rgb]{1,0.97,0.97} & 0.21 \cellcolor[rgb]{1,0.93,0.93} & 0.96 \cellcolor[rgb]{1,0.92,0.92} & 0.03 \cellcolor[rgb]{1,0.94,0.94} & 0.16 \cellcolor[rgb]{1,0.92,0.92} \\
 \textit{No Motion Cost} & 5.24 \cellcolor[rgb]{1,0.95,0.95} & 0.16 \cellcolor[rgb]{1,0.98,0.98} & 0.48 \cellcolor[rgb]{1,0.84,0.84} & 2.82 \cellcolor[rgb]{1,0.76,0.76} & 0.08 \cellcolor[rgb]{1,0.85,0.85} & 0.49 \cellcolor[rgb]{1,0.75,0.75} \\
 \textit{Motion Cost Planner} & 25.44 \cellcolor[rgb]{1,0.75,0.75} & 0.86 \cellcolor[rgb]{1,0.91,0.91} & 0.41 \cellcolor[rgb]{1,0.86,0.86} & 1.92 \cellcolor[rgb]{1,0.84,0.84} & 0.06 \cellcolor[rgb]{1,0.88,0.88} & 0.32 \cellcolor[rgb]{1,0.84,0.84} \\
 \textit{Exploration Path w/ Cost Optimizer} & 37.17 \cellcolor[rgb]{1,0.63,0.63} & 1.70 \cellcolor[rgb]{1,0.83,0.83} & 0.34 \cellcolor[rgb]{1,0.89,0.89} & 1.44 \cellcolor[rgb]{1,0.88,0.88} & 0.05 \cellcolor[rgb]{1,0.91,0.91} & 0.24 \cellcolor[rgb]{1,0.88,0.88}
\end{tabular}
\caption{The collision rate, \edited{motion cost and motion risk} for all compared methods over all robots. \edited{For the path collision rate,} the left column indicates the percentage of paths where any pose was in collision. The right column shows the percentage of paths where over $\frac{1}{3}$ of all poses were in collision. This serves as an indicator of how many paths would produce severe collisions which could become dangerous for the robot. \edited{For motion cost and motion risk, the columns indicate their mean and 95th-percentile.}}
\label{tab:collision_rate}
\end{table}

\Cref{tab:collision_rate} shows the collision rates computed for each planner.
In addition to computing the general collision rate of each planner, which is the ratio of paths which have any collision, we also computed a severe collision rate.
We did this to diminish the effect of slight path collisions, which likely would not be problematic for the robot, or are caused by map artifacts or state estimation drift.
A path is in severe collision if more than $\frac{1}{3}$ of its poses are in collision.
For this, we considered only paths with at least three poses, to avoid the outsized impact short paths would have on this metric.

All reachability planners performed at safe levels w.r.t path collisions.
Interestingly, ArtPlanner deployed during the competition performed even better than during playback.
\edited{This is most likely related to differences in message timing between recorded competition data and playback, but not the actual underlying performance.}
Note that collision rates are not zero because we checked collisions against height maps at the time paths are published.
This means, map changes, artifacts and state estimation drift can lead to some map collisions which are unproblematic during operation.
As mentioned before, we only had a single real collision during the event, and in that case the obstacle was missing from the map (see \Cref{sec:results_path_follower}).
The \textit{Exploration Path} only has a slightly higher general collision rate at $10.93\%$ but the highest rate of severe collisions at $2.44\%$.
This is due to the fact that the exploration planner performs collision checking in a coarser, volumetric map.
This map captures most larger obstacles, which leads to a low general collision rate.
Small obstacles, however, can be completely missed, which leads to severe collisions in these instances.
\textit{Motion Cost Planner} and \textit{Exploration Path w/ Cost Optimizer} have very high general collision rates at $25.44\%$ and $37.17\%$, respectively, but comparatively much lower severe collision rates, at $0.86\%$ and $1.70\%$.
This is due to the fact that, when training the cost network, collisions with the environment are allowed, as long as the robot still reaches its intended goal.
This means paths frequently slightly graze some obstacles without severely colliding.
While mostly not fatal, in practice even slight bumps into obstacles should be avoided since they can get the robot stuck on protruding elements or damage its payload.
The higher severe collision rate of the \textit{Exploration Path w/ Cost Optimizer} was caused by the colliding initial exploration path, which the gradient based cost optimizer could not fix.

Taking a look at the collision heat maps in \Cref{fig:cost_maps} allows us to determine in which situations these planners struggled.
\Cref{fig:cost_maps}~\mynumber{2} shows an instance where the \textit{Exploration Path} passes through obstructing traffic cones.
Note that this does not show up as a severe hot spot on the heat map due to \textit{Exploration Path} pruning for collision checking, as discussed in \Cref{sec:results_methodology}.
The \textit{Motion Cost Planner} generally had few collisions, except in a narrow cave section, where the optimized path performs a turn due to bad initialization (\Cref{fig:cost_maps} \mynumber{5}).
The \textit{Exploration Path w/ Cost Optimizer} collisions were generally caused by the optimizer not dealing well with height map artifacts (\Cref{fig:cost_maps} \mynumber{6}+\mynumber{7}).

\subsubsection{Planning Time}

\begin{figure}
  \includegraphics[width=\linewidth]{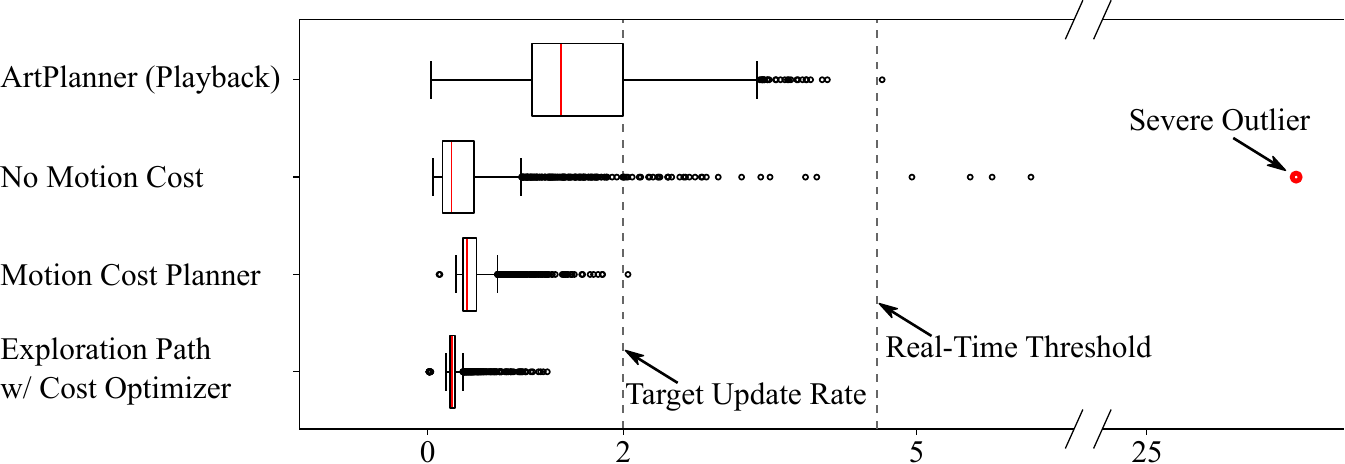}
  \caption{While ArtPlanner plans slower than the compared methods on average, its computation time is bounded. This avoids severe planning time outliers the \textit{No Motion Cost} planner produces in rare cases. The \textit{Motion Cost Planner} is fast because it checks a fixed planning graph pattern and therefore does not perform any pose sampling. The \textit{Cost Optimizer} only does a few steps of gradient-descent which makes it the fastest method in our comparison.}
  \label{fig:planning_times}
\end{figure}

Since we did not log planning times of our planner during the competition, and GBPlanner2 has a different scope to the other navigation planners, we compare the planning times of all methods during playback.
Planning times are shown as box plots in \Cref{fig:planning_times}.
Our chosen target update rate was to publish a new path every 2 seconds and the real-time threshold for our \SI{8}{\meter}$\times$\SI{8}{\meter} map at a locomotion speed of \SI{0.9}{\meter\per\second} was \SI{4.44}{\second}.
The real-time threshold~\cite{wellhausen2021rough} is the time the robot requires to reach the edge of the height map at maximal speed.

We set ArtPlanner's maximal sampling time $T$ to 2 seconds, which does not factor in map processing and motion cost query time.
Therefore, ArtPlanner can exceed the target time if the maximum sampling time is reached, with a maximal planning time of \SI{4.65}{\second}.
Consequently, we achieved the target time in 75\% of cases, exceeding the real-time threshold only a single time, by \SI{0.21}{\second}.
The \textit{No Motion Cost} planner has fast query times in the median.
Note that these times do not include the graph sampling time, because this method uses a persistent graph which is built over time, in-between planning queries.
If we added this sampling time the median performance would be comparable to ArtPlanner.
Unfortunately, this method can produce severe planning time outliers on rare occasion due to the graph validation at query time, as discussed in \Cref{sec:method_graph_construction}.
We observed a time of \SI{26.53}{\second} in the data we used to generate \Cref{fig:planning_times} but observed even longer times in other instances.
This shows the benefit of our graph validation method through batch motion cost query.
We can limit our planning time and can therefore avoid the planner being unresponsive for a long period of time.
The \textit{Motion Cost Planner} uses a fixed planning graph and therefore only has to perform a batch motion cost query, but no sampling.
Since the optimization stage also always performs a fixed number of iterations, planning is fast and the target time can be reached in all cases.
Since the \textit{Exploration Path w/ Cost Optimizer} only deploys the optimization stage of the \textit{Motion Cost Planner} it is even faster.

Overall, all evaluated planners would be fast enough for real-time operation in the nominal case.
However, the worst-case time of the \textit{No Motion Cost} planner is too long to deploy it in a competition environment.

\subsection{Component Analysis}

We now investigate the effect individual components of our method had on the performance during the \textit{Finals}.

\subsubsection{\edited{Motion Cost Analysis}}

\begin{figure}
  \begin{subfigure}[t]{0.49\linewidth}
    \centering
    \includegraphics[width=\linewidth]{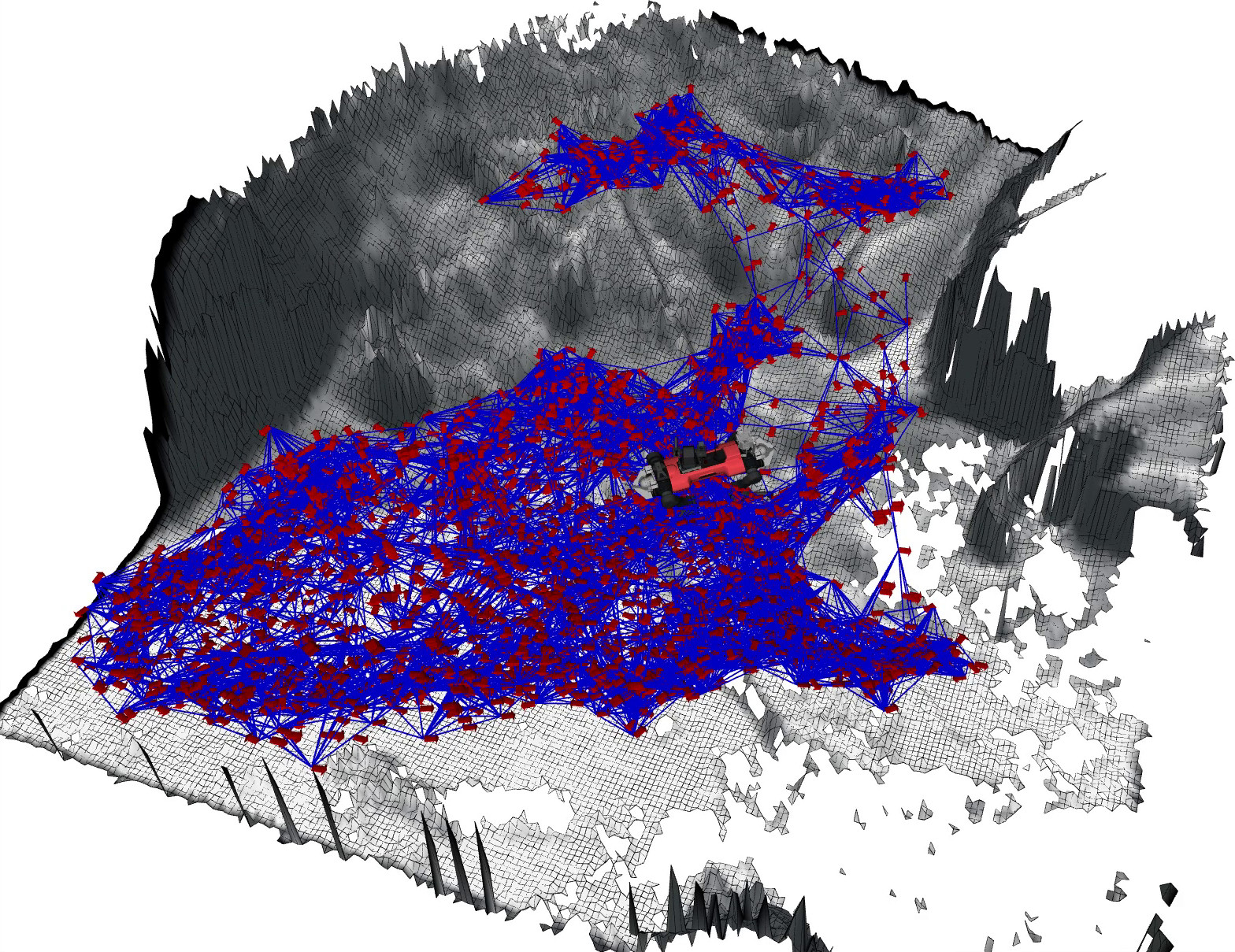}
    \caption{Without risk pruning}
  \end{subfigure}
  \hfill
  \begin{subfigure}[t]{0.49\linewidth}
    \centering
    \includegraphics[width=\linewidth]{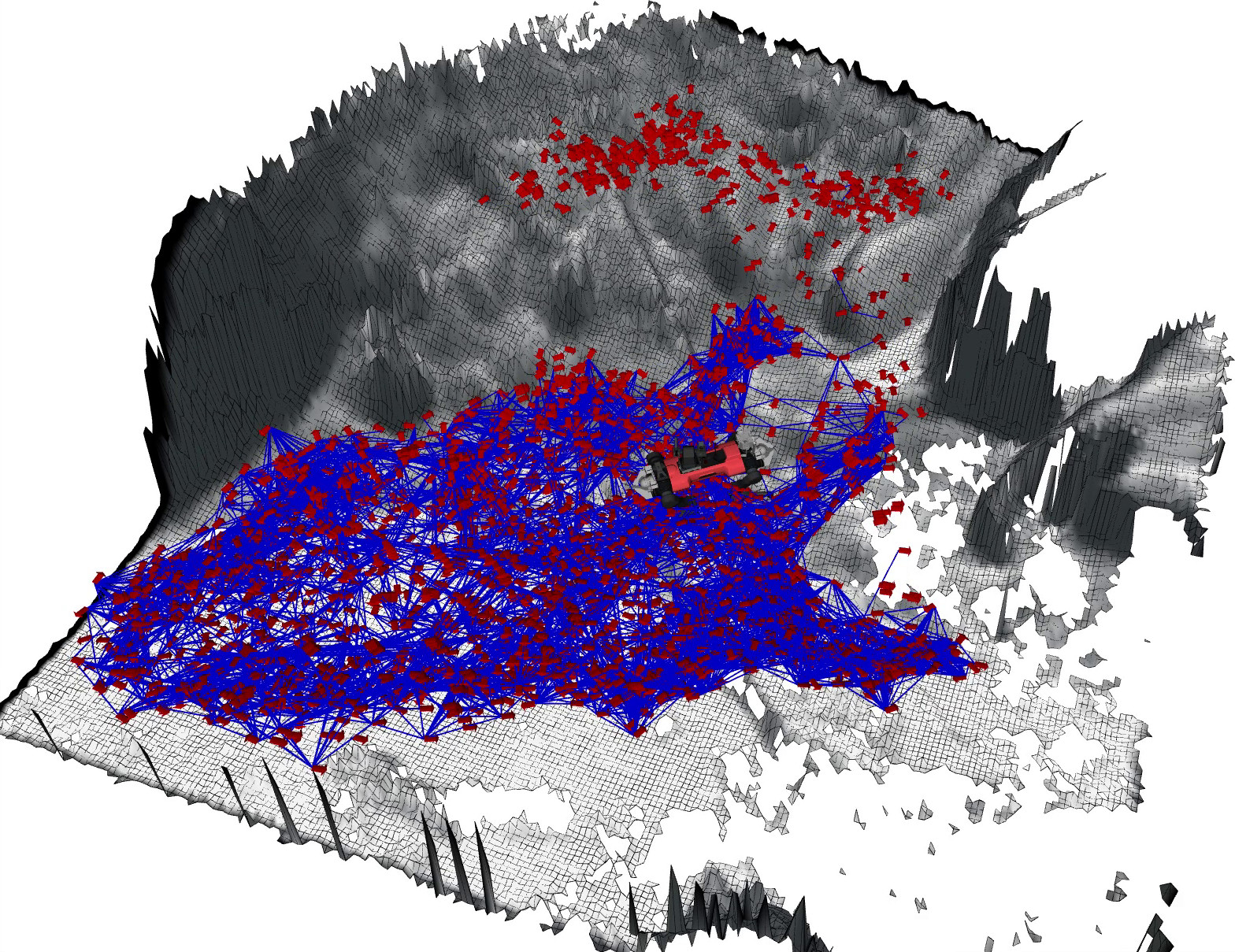}
    \caption{With risk pruning}
  \end{subfigure}
  \caption{Sampled valid poses shown in red with blue graph edges connecting them. Using reachability checking, valid robot poses are still found on a steep and rocky incline in the cave section. Pruning graph edges based on motion risk prevents the planning graph from spanning this risky area.}
  \label{fig:graph_validation}
\end{figure}

We used the motion cost in two ways, to prune the planning graph based on motion risk, and to optimize the cost function for both risk and time.

Since the terrain during the \textit{Finals} was generally flat and easy to traverse for our locomotion controller~\cite{miki2022learning} in most sections, risk pruning did not have a large effect. 
We observed the biggest effect in the large hall of the Cave section, which housed a rocky incline (bottom right of map in \Cref{fig:cost_maps}). 
Since the inclination of the terrain was in principle low enough to climb it, GBPlanner2 wanted to explore up the slope, even though the actual terrain was too rough and rocky to overcome.
Using just reachability checking, our sampler found individual valid poses on the slope and connected them, as shown in \Cref{fig:graph_validation}(a).
Here, risk pruning identified that moving on this slope was too risky and removed these edges, as shown in \Cref{fig:graph_validation}(b).
Without risk pruning the robot would have tried to scale this slope, which would in the best case have lead to lost time, and in the worst case to loss of this robot.

\Cref{fig:cost_maps}~\mynumber{3}+\mynumber{5} show that the cost function generally lead to safer paths, which kept a safe distance from obstacles.
This can be attributed to the risk term $c_r$ in the cost function.
The time cost $c_t$ had negligible impact compared to the shortest path of the \textit{No Motion Cost} planner, since the terrain present during the \textit{Finals} was uniform enough such that the shortest path generally was also the fastest. 
However, $c_t$ was necessary to condition the planning problem, since a pure risk cost would have lead to large detours to achieve minor risk improvements.

\subsubsection{\edited{Safety Threshold Analysis}}
\label{sec:results_safety_threshold}

\begin{figure}
  \includegraphics[width=\linewidth]{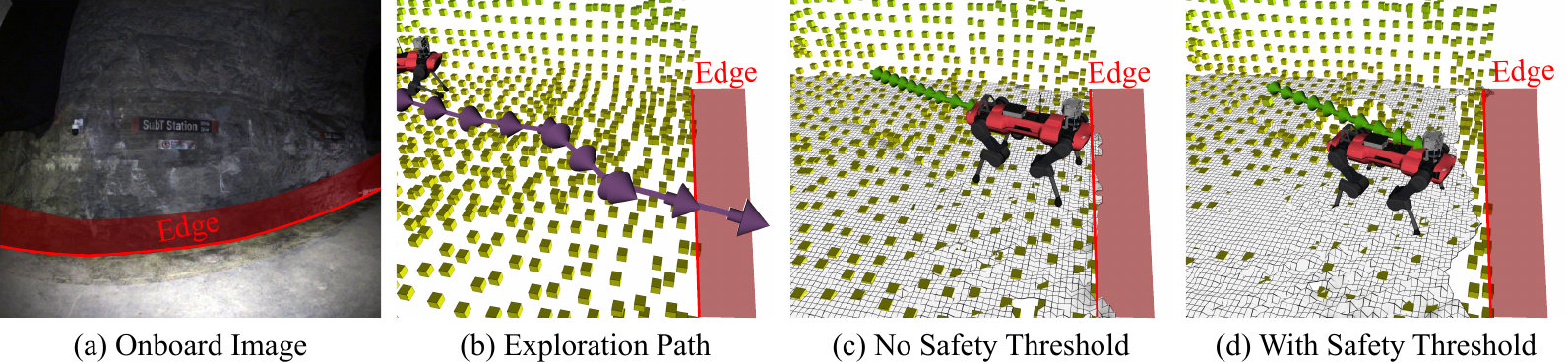}
  \caption{(a) The SubT Station platform had a sharp edge with a significant drop. (b) GBPlanner2 was tuned to be optimistic and planned over the edge of the SubT Station platform. (c) Without a foothold safety margin the robot would have stepped onto and possibly over the platform edge. (d) With foothold safety margin the final path pose is a safe distance from the platform edge. Yellow cubes represent our lidar map to indicate the actual location of the edge. The ANYmal model in (c)+(d) does not indicate the current robot pose, but rather is placed at the final pose of the path to better show how close the robot would have stepped to the edge.}
  \label{fig:safety_threshold}
\end{figure}

The safety threshold was introduced to handle negative obstacles.
One carrier robot reached the Subway Station in autonomous exploration mode during the first Preliminary Run of the \textit{Finals}. 
The Subway Station had a sharp drop with a wall a few meters behind, pictured in \Cref{fig:safety_threshold}(a).
As discussed in \Cref{sec:intro_navigation_stack}, the exploration planner was tuned to be optimistic, and planned to explore into the free space above the train tracks (\Cref{fig:safety_threshold}(b)).
With the wall visible behind the platform, both image inpainting and virtual surfaces would have simply created flat ground or a gentle slope such that we could not rely on motion cost for safety.
Reachability checking generally prevents the planner from planning over the edge, however, without a safety threshold, the planned final pose is dangerously close to the edge (\Cref{fig:safety_threshold}(c)).
With the safety threshold applied, the robot only plans to a safe distance from the edge (\Cref{fig:safety_threshold}(d)).

The safety threshold therefore most likely prevented a severe fall of the robot during the first Preliminary Run, which would have caused heavy damage to the payload on top of the robot and possibly the robot itself.

The safety margin parameters were well tuned and generally did not cause the robot to be overly cautious.
We only observed the safety margin to come into effect around the tall railroad tracks shown in \Cref{fig:cost_maps}~\mynumber{1}.
These were just on the edge of being traversable and tall enough to cause occlusions in the height map.
This effected a few unsmooth, but still safe paths when the exploration path crossed these tracks, but mostly caused our planner to prefer staying in between the rail tracks, which was the safest approach, anyway.

\subsubsection{\edited{Virtual Surfaces Analysis}}
\label{sec:results_virtual_surfaces}

\begin{figure}
  \begin{subfigure}[t]{0.32\linewidth}
    \centering
    \includegraphics[width=\linewidth]{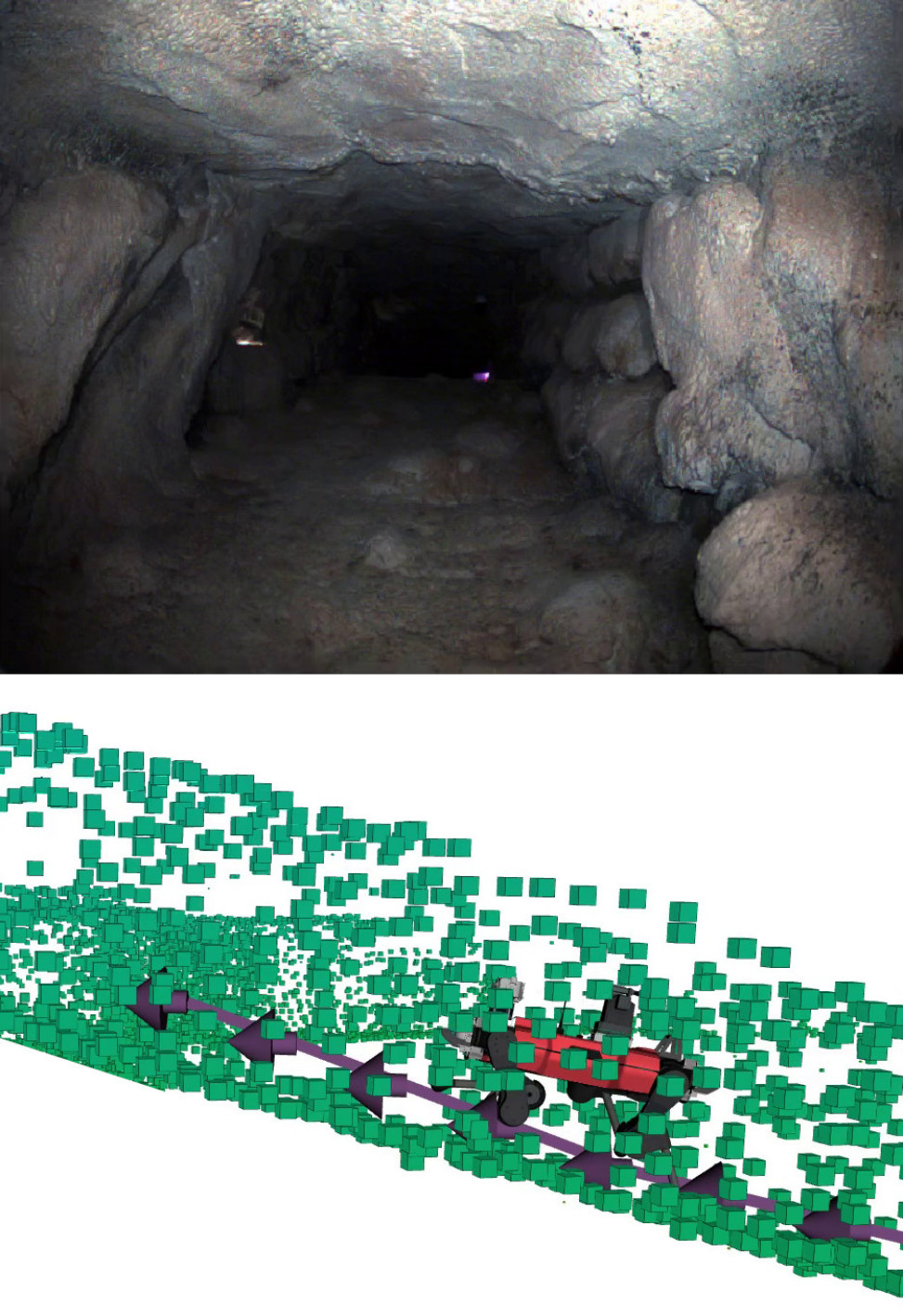}
    \caption{Onboard image and Lidar map.}
  \end{subfigure}
  \hfill
  \begin{subfigure}[t]{0.18\linewidth}
    \centering
    \includegraphics[width=\linewidth]{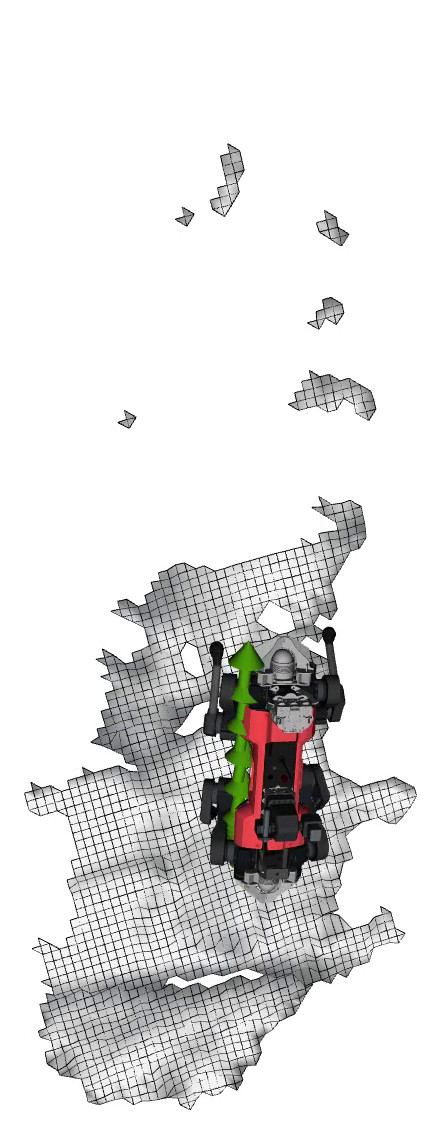}
    \caption{Steppable region with height map only.}
  \end{subfigure}
  \hfill
  \begin{subfigure}[t]{0.18\linewidth}
    \centering
    \includegraphics[width=\linewidth]{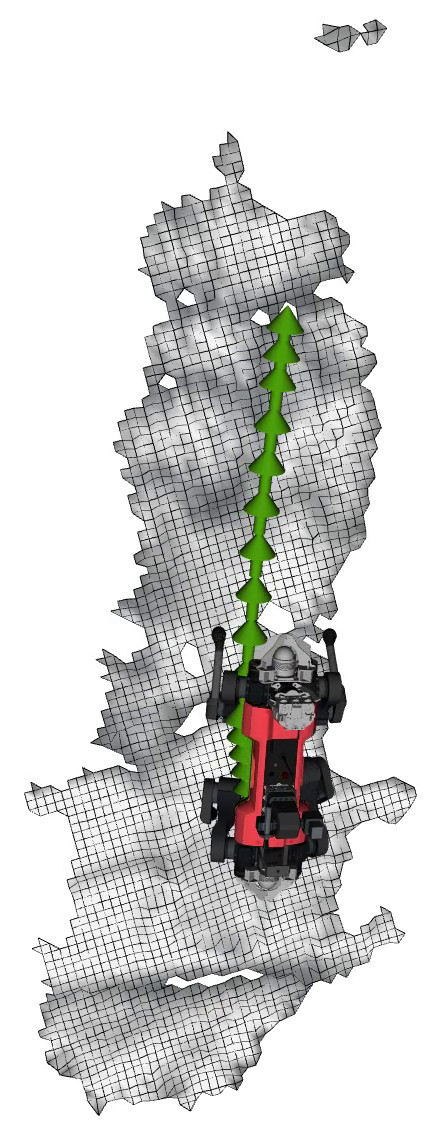}
    \caption{Steppable region with virtual surfaces.}
  \end{subfigure}
  \caption{(a) The cave section had a narrow incline leading up to a Cube artifact. (b) Due to the low sensor height above the uneven ground, the height map is too sparse to plan uphill. (c) Using virtual surfaces, the plannable area is vastly increased.}
  \label{fig:upper_bound}
\end{figure}

As discussed in \Cref{sec:method_virtual_surfaces}, we only use virtual surfaces above sensor height, due to safety concerns with negative obstacles, demonstrated in \Cref{sec:results_safety_threshold}.
Since the \textit{Finals} course was mostly planar, and did not have multiple levels like the Urban Circuit, the only time virtual surfaces came into effect were in the cave section (located just to the left of \Cref{fig:cost_maps} \mynumber{5}).
Here, the course had a two-sided ramp with a very low ceiling height, which lead to a platform with a Cube artifact, pictured in \Cref{fig:upper_bound}(a).
Due to the rough terrain on the incline, the height map contained many holes in front of the robot.
This reduced the size of the steppable area so much that planning would have become almost impossible, as shown in \Cref{fig:upper_bound}(b).
Here, \Cref{fig:upper_bound}(c) shows how the virtual surfaces allowed us to plan up the incline, reach the platform, and score the cube artifact.

\subsection{Issues}

Although our planner performed very well during the \textit{Finals}, it still encountered some issues related to other parts of the navigation stack.

\subsubsection{Height Map Spikes}
\label{sec:results_map_spikes}

\begin{figure}
  \includegraphics[width=\linewidth]{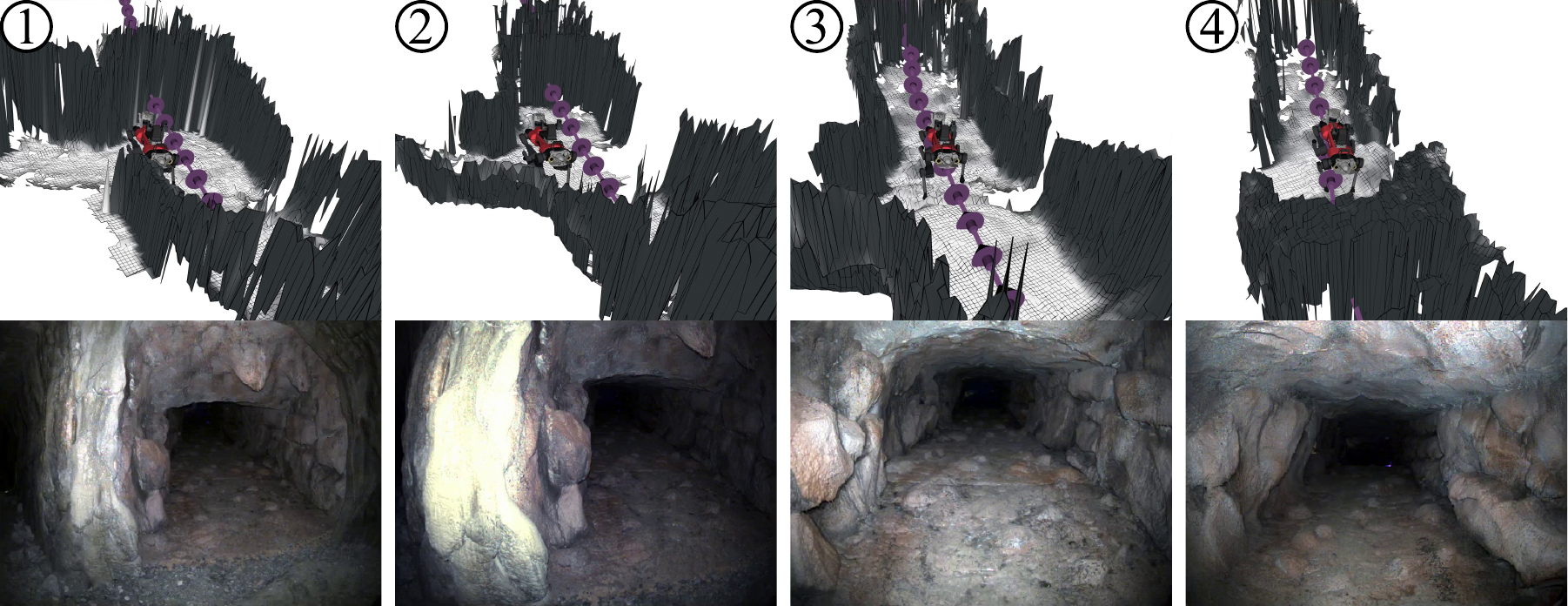}
  \caption{Corrupted height maps from ceiling lidar returns frequently slowed down progress by reducing the distance the planner can plan ahead. \mynumber{1} The exploration path leads into a low corridor which is observed as a wall in the height map. \mynumber{2} The fake wall shifts forward as the robot gets closer. \mynumber{3} The incline reduces ceiling hits in front of the robot which helps move the fake wall farther from the robot. \mynumber{4} Once fully on the slope, ceiling measurements immediately behind the robot produce another fake wall.}
  \label{fig:map_spikes}
\end{figure}

As mentioned in \Cref{sec:method_ceiling_point_filter}, obtaining a clean height map in environments with low ceilings was challenging, and we tuned our ceiling point filter to also work with inclines and stairs, which exacerbated the issue.
This slowed our progress quite a bit in the cave section, where the ceiling was especially low.
Unfortunately, we reached these sections with our explorer robots, which recorded many ceiling points very close to the robot, due to their dome lidar configuration (see \Cref{fig:anymal_hardware}(a)+(b)).

Although this slowed our progress, we never got stuck.
We demonstrate this behavior with a narrow cave opening, which is immediately followed by an incline, shown in \Cref{fig:map_spikes}.
Even when the robot was less than a meter from the opening (\Cref{fig:map_spikes}~\mynumber{1}), the height map showed a straight wall.
Our planner could therefore only plan right up to the halucinated wall, which then shifts forward slightly (\Cref{fig:map_spikes} \mynumber{2}) and allows the robot to plan a bit farther.
In \Cref{fig:map_spikes} \mynumber{3}, the slope aligns with the distance-dependent height threshold of our ceiling point filter and we can plan forward farther.
Once we are fully on the incline (\Cref{fig:map_spikes} \mynumber{4}), the negative slope at the rear of the robot causes the fake wall to reappear right behind it.
Nonetheless, the robot was able to plan up the slope and back down fully autonomous.

\subsubsection{Path Follower}
\label{sec:results_path_follower}

\begin{figure}
  \includegraphics[width=\linewidth]{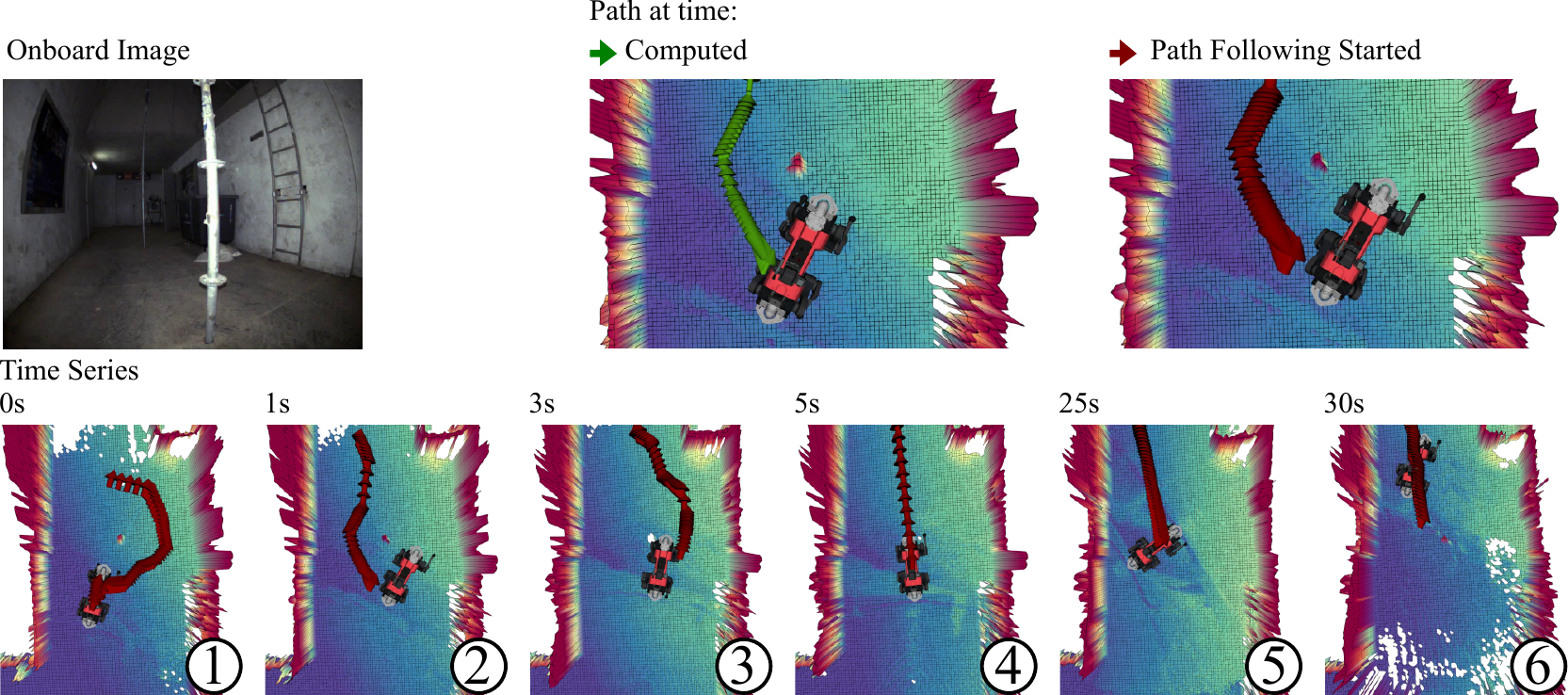}
  \caption{A \SI{500}{\milli\second} time delay between computing a path and starting to follow it lead to imperfect path following. As a result, one robot briefly got stuck on a narrow scaffolding pole in the urban section. \mynumber{1} The robot initially planned to pass the pole on the right but \mynumber{2} replanned to use the shorter path on the left. \mynumber{3} Since the robot already moved, the next path went right again which got the robot stuck on the pole. \mynumber{4} The pole got removed from the map as it was now too close to be perceived and, consequently, paths through the pole were planned until \mynumber{5} the robot drifted to the left and \mynumber{6} finally got unstuck.}
  \label{fig:path_follower}
\end{figure}

Planners rely on their paths being tracked accurately.
The path follower is therefore an important component of the navigation stack.
Our pure-pursuit path follower generally performed well, tracking paths precisely in narrow spaces and moving swiftly in more open spaces.
However, during data analysis for this work, we discovered that there was a non-constant delay between the planner publishing a path, and the path follower starting to track it, which we observed to be up to \SI{500}{\milli\second}.
The top right of \Cref{fig:path_follower} shows an example of the difference in robot position between the plan being published and being followed.
This, in combination with height map issues and the unsmooth nature of sampling-based planning, caused one explorer to get stuck on a scaffolding pole in the Urban section for about 20 seconds.
The time series of this event is shown at the bottom of \Cref{fig:path_follower}:

\mynumber{1} When the robot approaches the pole, the initial plan is to circumnavigate it on the right, which is slightly longer than on the left. 
\mynumber{2} Just as the robot starts moving, a new path is published, which passes the pole on the left.
\mynumber{3} This causes the robot to move so close to the pole that it cannot be perceived and disappears from the height map.
Since the robot had followed the original path to the right, the next path goes right again and gets the robot stuck on the pole.
\mynumber{4} Because the pole is now missing from the height map, all new paths go straight through it.
\mynumber{5} After another 20 seconds, the robot has drifted to the left of the pole and \mynumber{6} finally gets unstuck.

This event demonstrates the complexity induced by interactions of the many components which make up a robotic system.
While every path planned before the pole disappeared from the map was safe, this was no guarantee that the actual robot motion was safe.
In the end, the robot only continued due to our exceptionally robust locomotion controller~\cite{miki2022learning}.


\section{Conclusion}
\label{sec:conclusion}

In this work we presented ArtPlanner, a navigation planner for legged robots, which was deployed by team CERBERUS during the \acf{SubT} \textit{Finals} to win the competition.
Its reachability robot abstraction allows for precise maneuvering in tight spaces, while its learned motion cost makes sure the path is safe and cost-optimal.
We presented extensive real-world evaluations of the planner performance during the \ac{SubT} \textit{Finals} and compared to other state-of-the-art navigation planners.
We showed that ArtPlanner outperforms the other methods and that without ArtPlanner we likely would have encountered navigation issues during the competition.
Our planner never failed to plan and never produced an infeasible or unsafe path and all deployed ground robots remained operational throughout the competition.

\subsection{Limitations}

We build a new planning graph every time the planning map is updated and only allow for a limited sampling budget.
This means that the algorithm proposed in this work is not probabilistically complete.
This is a limitation of this work but has shown to be unproblematic in practice, since ArtPlanner runs fast enough to sample its fixed-size map densely, while maintaining real-time update rates.
Batch risk querying for edge validation is compatible with our previously used persistent graph~\cite{wellhausen2021rough}, which is probabilistically complete.
However, the error-prone and potentially expensive accounting for updated graph edges would still be necessary.
We found this trade-off of theoretical guarantees for practical reliability to be a good choice for our use-case.

\subsection{Future Work}

The main issues encountered during the \ac{SubT} \textit{Finals} were related to the 2.5D height map representation and the interaction with the path follower.
Moving to a full 3D representation would be desirable, however, until now these methods could not run in real-time with the level of detail necessary for legged robot navigation.
Increased computational power of edge computing devices and advanced GPU-accelerated algorithms could soon enable real-time mapping at necessary resolutions on mobile robots.
Another direction for research is to make the exploration planner itself robot-traversability aware. 
Recently proposed 3D motion cost networks~\cite{frey2022locomotion} could prevent instances where the exploration planner proposes a path which the robot cannot follow.
Finally, \ac{RL} has the potential to remove the limitations inherently produced by any map representation and can at the same time replace the path follower.
Recent results in sim-to-real training of \ac{RL}-policies show the potential for training a navigation agent in simulation, which would allow us to go directly from sensor measurements to locomotion command.

\subsection{Acknowledgements}

We would like to thank the following people for their help in the context of this work.
Samuel Zimmermann for writing the code which interfaced our planner with the rest of the ANYmal software stack.
Samuel Zimmermann, Marco Tranzatto, Gabriel Waibel and Timon Homberger for frequent testing in the run-up to the \textit{Finals}.
Takahiro Miki for implementing the ceiling point filter.
Marko Bjelonic for helpful feedback on the manuscript.
The entire team CERBERUS for exceptional performance over the four years of the competition.




\bibliographystyle{apalike}
\bibliography{references}

\end{document}